\documentclass{article}

\usepackage{microtype}
\usepackage{graphicx}
\usepackage{subfigure}
\usepackage{booktabs} 
\usepackage{chngcntr}
\usepackage{amsmath}
\usepackage{xcolor}
\usepackage{amssymb}

\usepackage{hyperref}

\usepackage[nohyperref,accepted]{icml2019}

\def \papertitle{Structured agents for physical construction}
\icmltitlerunning{\papertitle}

\DeclareMathOperator*{\argmax}{arg\,max}

\newcommand{\silhouette}[0]{\emph{Silhouette}}
\newcommand{\reaching}[0]{\emph{Connecting}}
\newcommand{\coveringone}[0]{\emph{Covering Hard}}
\newcommand{\coveringtwo}[0]{\emph{Covering}}
\newcommand{\supplemental}[0]{Supplemental}

\newcommand{\silhouetteabsoluterszerogn}{\href{https://drive.google.com/file/d/1ADAIWzZLZLqMbBrH-FXn_kwnlFCZfIlY/view?usp=sharing}{
GN-RS0}}

\newcommand{\reachingabsoluterszerornn}{\href{https://drive.google.com/file/d/1sGAHkH3MKjfFJHDlE0wQ0F2BVVqrGtte/view?usp=sharing}{
RNN-RS0}}
\newcommand{\reachinggenoneabsoluterszerornn}{\href{https://drive.google.com/file/d/12EBMDrT2tBg9ixjubRCPtYuq9WUnZcAV/view?usp=sharing}{
RNN-RS0}}
\newcommand{\reachinggentwoabsoluterszerornn}{\href{https://drive.google.com/file/d/1N8TSPz_Gnby0pEnWzdFDeQaN-l3HyppV/view?usp=sharing}{
RNN-RS0}}
\newcommand{\coveringoneabsoluterszerogn}{\href{https://drive.google.com/file/d/1_c7dEBwQSZzfi4wu78z2XuAKY1GlXp3C/view?usp=sharing}{
GN-RS0}}
\newcommand{\coveringtwoabsoluterszerogn}{\href{https://drive.google.com/file/d/1_qXySg4Tv8sjOyO2GA6yPQTtQA35gd5w/view?usp=sharing}{
GN-RS0}}

\newcommand{\silhouetterelativerszerornn}{\href{https://drive.google.com/file/d/1xBMTUBA7uXd1JzMMdauxA4hCfsELxq-E/view?usp=sharing}{
RNN-RS0}}

\newcommand{\silhouettegenonerelativerszerocnn}{\href{https://drive.google.com/file/d/1KUlFyiKM-6AR7n5drwtKvAvcedMy42Uc/view?usp=sharing}{
CNN-RS0}}
\newcommand{\reachingrelativerszerornn}{\href{https://drive.google.com/file/d/1R234HRwUsIwDmK8LdnMGdugmslHDeip3/view?usp=sharing}{
RNN-RS0}}
\newcommand{\reachinggenonerelativerszerornn}{\href{https://drive.google.com/file/d/1O_eeVQsMIWCqYEczav7r7Bj3OCjq3YI7/view?usp=sharing}{
RNN-RS0}}
\newcommand{\reachinggentworelativerszerornn}{\href{https://drive.google.com/file/d/1htPzkztSSBaLLgZMreCixtUPKgQhEF_A/view?usp=sharing}{
RNN-RS0}}
\newcommand{\coveringonerelativerszerornn}{\href{https://drive.google.com/file/d/1dxFWPeOSWsHWBx-457xNVLXM92gPy42f/view?usp=sharing}{
RNN-RS0}}
\newcommand{\coveringtworelativerszerocnn}{\href{https://drive.google.com/file/d/1C2-3rEVXWwBhMrYHD6xc-dqTl8yAxFFf/view?usp=sharing}{
CNN-RS0}}

\newcommand{\silhouetterelativedqngn}{\href{https://drive.google.com/file/d/1Apovgr2YT6S3hwmvP4qExTCAcMWEmDpx/view?usp=sharing}{
GN-DQN}}
\newcommand{\silhouettegenonerelativedqngn}{\href{https://drive.google.com/file/d/1uQzxqRJ2qOogpG3SjfkKB36BHvVSg91a/view?usp=sharing}{
GN-DQN}}
\newcommand{\reachingrelativedqngn}{\href{https://drive.google.com/file/d/1CthEtlLN3Bkee4kB2N4dn36P8pYMppQH/view?usp=sharing}{
GN-DQN}}
\newcommand{\reachinggenonerelativedqngn}{\href{https://drive.google.com/file/d/1dAeTBw7aLVpYi4M-IpLwsZSfbBATlKEs/view?usp=sharing}{
GN-DQN}}
\newcommand{\reachinggentworelativedqngn}{\href{https://drive.google.com/file/d/1hQzWjF21P8PEqShV_OBNwjVZ9fQ4CEfF/view?usp=sharing}{
GN-DQN}}
\newcommand{\coveringonerelativedqngn}{\href{https://drive.google.com/file/d/1FWDDqb73qCtWARlDNhO-CzM4k1L7P5fc/view?usp=sharing}{
GN-DQN}}
\newcommand{\coveringtworelativedqngn}{\href{https://drive.google.com/file/d/18_dHa3_v0iJB6boYkBmXOBpzqVCsWe8F/view?usp=sharing}{
GN-DQN}}

\newcommand{\silhouetterelativedqngnmcts}{\href{https://drive.google.com/file/d/14-3FjlWYBMf184TOfj5lSvVr2sZau95a/view?usp=sharing}{
GN-DQN with MCTS at test time}}
\newcommand{\silhouettegenonerelativedqngnmcts}{\href{https://drive.google.com/file/d/14UBsB7K2opC7IuvSQ3SQivgoE9BL6R8B/view?usp=sharing}{
GN-DQN with MCTS at test time}}
\newcommand{\reachingrelativedqngnmcts}{\href{https://drive.google.com/file/d/1OorieFAptwQn0WvwbdUZb-2Czx2Jv7tz/view?usp=sharing}{
GN-DQN with MCTS at test time}}
\newcommand{\reachinggenonerelativedqngnmcts}{\href{https://drive.google.com/file/d/1RZ4sUQQDCxIkkINF7doDQ5_ehsHmIbBe/view?usp=sharing}{
GN-DQN with MCTS at test time}}
\newcommand{\reachinggentworelativedqngnmcts}{\href{https://drive.google.com/file/d/1-zzNtrF0_iA1i3uSfJ0WxmrRyZevsI1R/view?usp=sharing}{
GN-DQN with MCTS at test time}}

\newcommand{\coveringonerelativedqngnmctstrain}{\href{https://drive.google.com/file/d/1agmhvX1zcGKotXSQ7SWM92P4im747L6g/view?usp=sharing}{
GN-DQN with MCTS at train and test time}}
\newcommand{\coveringtworelativedqngnmcts}{\href{https://drive.google.com/file/d/1GffHlqXOzcJPKT3f3paqeFpce4VRM3L7/view?usp=sharing}{
GN-DQN with MCTS at test time}}

\definecolor{mediumgreen}{rgb}{0,.6,0}

\begin{document}

\twocolumn[
\icmltitle{Structured agents for physical construction}




\icmlsetsymbol{equal}{*}

\begin{icmlauthorlist}
\icmlauthor{Victor Bapst}{equal,dm}
\icmlauthor{Alvaro Sanchez-Gonzalez}{equal,dm}
\icmlauthor{Carl Doersch}{dm}
\icmlauthor{Kimberly L.~Stachenfeld}{dm}
\icmlauthor{Pushmeet Kohli}{dm}
\icmlauthor{Peter W.~Battaglia}{dm}
\icmlauthor{Jessica B.~Hamrick}{dm}
\end{icmlauthorlist}

\icmlaffiliation{dm}{DeepMind, London, UK}

\icmlcorrespondingauthor{Jessica Hamrick}{jhamrick@google.com}

\icmlkeywords{Machine Learning}

\vskip 0.3in
]

 
\printAffiliationsAndNotice{\icmlEqualContribution} 

\begin{abstract}
Physical construction---the ability to compose objects, subject to physical dynamics, to serve some function---is fundamental to human intelligence.
We introduce a suite of challenging physical construction tasks inspired by how children play with blocks, such as matching a target configuration, stacking blocks to connect objects together, and creating shelter-like structures over target objects.
We examine how a range of deep reinforcement learning agents fare on these challenges, and introduce several new approaches which provide superior performance.
Our results show that agents which use structured representations (e.g., objects and scene graphs) and structured policies (e.g., object-centric actions) outperform those which use less structured representations, and generalize better beyond their training when asked to reason about larger scenes.
Model-based agents which use Monte-Carlo Tree Search also outperform strictly model-free agents in our most challenging construction problems.
We conclude that approaches which combine structured representations and reasoning with powerful learning are a key path toward agents that possess rich intuitive physics, scene understanding, and planning.
\end{abstract}

\section{Introduction}

Humans are a ``construction species''---we build forts out of couch cushions as children, pyramids in our deserts, and space stations that orbit hundreds of kilometers above our heads. 
What abilities do artificial intelligence (AI) agents need to possess to perform such achievements?
This question frames the high-level purpose of this paper: to explore a range of tasks more complex than those typically studied in AI, and to develop approaches for learning to solve them.

\begin{figure*}[t!]
\begin{center}
    \includegraphics[width=0.99\textwidth]{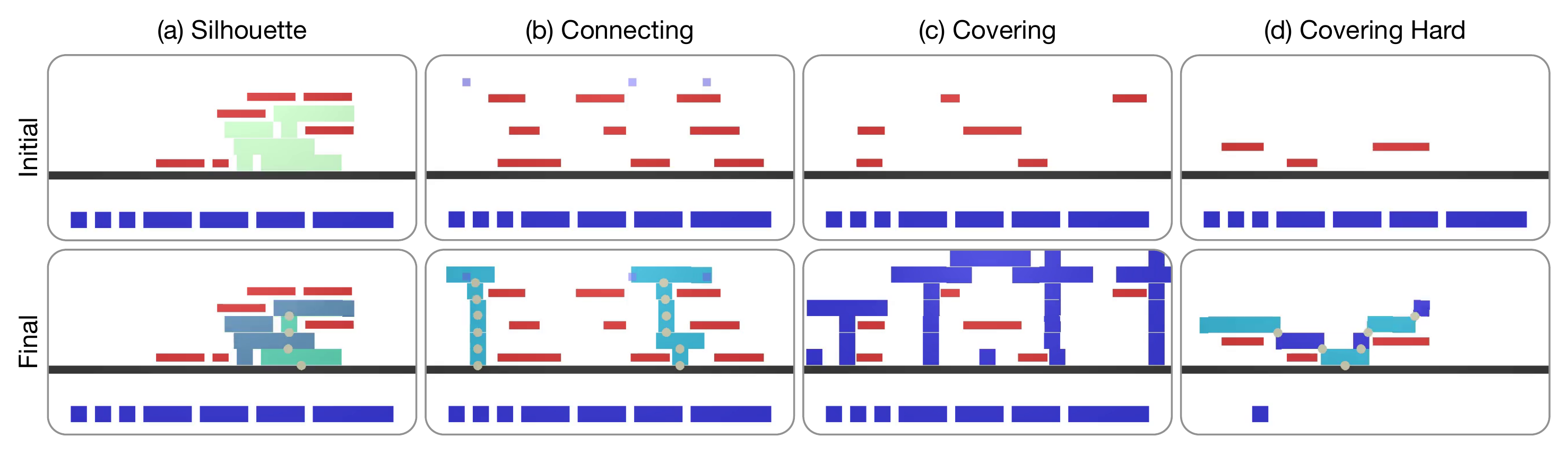}
    \caption{Construction task suite. In all tasks, dark blue objects are regular blocks, light blue blocks are ``sticky'', red objects are obstacles which cannot be touched, and grey circles indicate points where blocks have stuck together. The black line indicates the floor separating the scene above, which is subject to physics, from the blocks below, which can be picked up and placed. (a) \silhouette{} task. The agent stacks blocks to match the target blocks (depicted as light green blocks). (b) \reaching{} task. The agent stacks blocks to connect the small blue target objects to the floor. (c) \coveringtwo{} task. The agent stacks blocks to shelter the obstacles from above. (d) \coveringone{} task. Similar to \coveringtwo{}, but crucially the agent has a limited supply of movable blocks. Videos of agent behaviors in these tasks are available at \href{https://drive.google.com/drive/folders/1lC9rQTuKYe-XxY0KedK49ymJS0eyicqr?usp=sharing}{https://tinyurl.com/y7wtfen9} and in \supplemental{} Table~\ref{table:videos}.}
    \label{fig:tasks}
\end{center}
\end{figure*}

Physical construction involves composing multiple elements under physical dynamics and constraints to achieve rich functional objectives.
We introduce a suite of simulated physical construction tasks (Fig.~\ref{fig:tasks}), similar in spirit to how children play with toy blocks, which involve stacking and attaching together multiple blocks in configurations that satisfy functional objectives.
For example, one task 
requires stacking blocks around obstacles to connect target locations to the ground.
Another task 
requires building shelters to cover up target blocks and keep them dry in the rain.
These tasks are representative of real-world construction challenges: they emphasize problem-solving and functionality rather than simply replicating a given target configuration, reflecting the way human construction involves forethought and purpose.

Real-world physical construction assumes many forms and degrees of complexity, but a few basic skills are typically involved: spatial reasoning (e.g. concepts like ``empty'' vs ``occupied''), relational reasoning (e.g. concepts like ``next to'' or ``on top of''), knowledge of physics (e.g., predicting physical interactions among objects), and planning the allocation of resources to different parts of the structure.
Our simulated task environment (Fig.~\ref{fig:tasks}) is designed to exercise these skills, while still being simple enough to allow careful experimental control and tractable agent training. 

While classic AI studied physical reasoning extensively \cite{chen1990advances,pfalzgraf1997geometric}, construction has not been well-explored using modern learning-based approaches.
We draw on a number of techniques from modern AI, combining and extending them in novel ways to make them more applicable and effective for construction.
Our family of deep reinforcement learning (RL) agents can support: (1) vector, sequence, image, and graph-structured representations of scenes; (2) continuous and discrete actions, in absolute or object-centric coordinates; (3) model-free learning via deep Q-learning \cite{mnih2015human}, or actor-critic methods \citep{heess2015learning,munos2016safe}; and (4) planning via Monte-Carlo Tree Search (MCTS) \cite{coulom2006efficient}.

We find that graph-structured representations and reasoning, object-centric policies, and model-based planning are crucial for solving our most difficult tasks, outperforming standard approaches which combine unstructured representations with policies that take absolute actions.
Our results demonstrate the value of integrating rich structure and powerful learning approaches as a key path toward complex construction behavior.

\section{Related Work}

Physical reasoning has been of longstanding interest in AI.
Early work explored physical concepts with an emphasis on descriptions that generalize across diverse settings \cite{winston1970learning}. 
Geometric logical reasoning was a major topic in symbolic logic research \cite{chou1987mechanical, arnon1988geometric}, leading to geometric theorem-provers \cite{bouma1995geometric}, rule-based geometric constraint solvers for computer-aided design \cite{aldefeld1988variation, schreck2012geometric}, and logic-based optimization for open-ended objectives in robotics \cite{toussaint2015logic}.
Classic work often focused on rules and structured representations rather than learning because the sample complexity of learning was often prohibitive for contemporary computers.

Modern advances in learning-based approaches have opened new avenues for using vector and convolutional representations for physical reasoning \cite{wu2015galileo,wu2016physics,wu2017learning,mottaghi2016newtonian,fragkiadaki2016learning,finn2016unsupervised,agrawal2016learning,lerer2016learning,li2016fall,groth2018shapestacks,bhattacharyya2018long,ebert2018visual}. A common limitation, however, is that due to their relatively unstructured representations of space and objects, these approaches tend not to scale up to complex scenes, or generalize to scenes with different numbers of objects, etc.

Several recent studies have explored learning construction, including learning to stack blocks by placing them at predicted stable points \cite{li2017visual}, learning to attach blocks together to stabilize an unstable stack \cite{hamrick2018relational}, learning basic block-stacking by predicting shortest paths between current and goal states via a transition model \cite{zhang2018composable}, and learning object representations and coarse-grained physics models for stacking blocks \cite{janner2019reasoning}. Though promising, in these works the physical structures the agents construct are either very simple, or provided explicitly as an input rather than being \emph{designed} by the agent itself. A key open challenge, which this paper begins to address, is how to learn to design and build complex structures to satisfy rich functional objectives.

A main direction we explore is object-centric representations of the scene and agent's actions \cite{diuk2008object,scholz2014physics}, implemented with graph neural networks \cite{Scarselli2009,bronstein2017geometric,gilmer2017neural,battaglia2018relational}.
Within the domain of physical reasoning, graph neural networks have been used as forward models for predicting future states and images \cite{battaglia2016interaction,chang2017compositional,watters2017visual,vansteenkiste2018relational}, and can allow efficient learning and rich generalization.
These models have also begun to be incorporated into model-free and model-based RL, in domains such as combinatorial optimization, motor control, and game playing \cite{dai2017learning,kool2018attentionTSP,hamrick2018relational,wang2018nervenet,sanchezgonzalez2018graph,zambaldi2018deep}.
There are several novel aspects to our graph network policies beyond these existing works, including the use of multiple actions per edge and graphs that change size during an episode.

\section{Physical Construction Tasks}
\label{sec:tasks}

Our simulated task environment is a continuous, procedurally-generated 2D world implemented in Unity \cite{juliani2018unity} with the Box2D physics engine \cite{catto2013box2d}. Each episode contains unmoveable obstacles, target objects, and floor, plus movable rectangular blocks which can be picked up and placed. 

On each step of an episode, the agent chooses an available block (from below the floor), and places it in the scene (above the floor) by specifying its position.
In all but one task (\coveringone{}---see below), there is an unlimited supply of blocks of each size, so the same block can be picked up and placed multiple times.
The agent may also attach objects together by assigning the property of ``stickiness'' to the block it is placing.
Sticky objects form unbreakable, nearly rigid bonds with objects they contact.
In all but one task (\reaching{}) the agent pays a cost to make a block sticky.
After the agent places a block, the environment runs physics forward until all blocks come to rest.

An episode terminates when: (1) a movable block makes contact with an obstacle, either because it is placed in an overlapping location, or because they collide under physical dynamics; (2) a maximum number of actions is exceeded; or (3) the task-specific termination criterion is achieved (described below). The episode always yields zero reward when a movable block makes contact with an obstacle. 

\textbf{\silhouette{} task} (Fig.~\ref{fig:tasks}a).
The agent must place blocks to overlap with target blocks in the scene, while avoiding randomly positioned obstacles. The reward function is: $+1$ for each placed block which overlaps at least $90\%$ with a target block of the same size; and $-0.5$ for each block set as sticky. The task-specific termination criterion is achieved when there is at least $90\%$ overlap with all targets.

This is similar to the task in~\citet{janner2019reasoning}, and challenges agents to reason about physical support of complex arrangements of objects and to select, position, and attach sequences of objects accordingly.
However, by fully specifying the target configuration, \silhouette{} does not require the agent to design a structure to satisfy a functional objective, which is an important component of our other tasks.

\textbf{\reaching{} task}
(Fig.~\ref{fig:tasks}b). The agent must stack blocks to connect the floor to three different target locations, avoiding randomly positioned obstacles arranged in layers. The reward function is: $+1$ for each target whose center is touched by at least one block, and $0$ (no penalty) for each block set to sticky. The task-specific termination criterion is achieved when all targets are connected to the floor.

By not fully specifying the target configuration, the \reaching{} task requires the agent to \emph{design} a structure with a basic function---connecting targets to the floor---rather than simply implementing it as in the \silhouette{} task.
A wider variety of structures could achieve success in \reaching{} than \silhouette{}, and the solution space is much larger because the task is tailored so that solutions usually require many more blocks.

\textbf{\coveringtwo{} task}
(Fig.~\ref{fig:tasks}c). The agent must build a shelter that covers all obstacles from above, without touching them. The reward function is: $+L$, where $L$ is the sum of the lengths of the top surfaces of the obstacles which are sheltered by blocks placed by the agent; and
$-2$ for each block set as sticky.
The task-specific termination criterion is achieved when at least $99\%$ of the summed obstacle surfaces are covered. The layers of obstacles are well-separated vertically so that the agent can build structures between them.

The \coveringtwo{} task requires richer reasoning about function than the previous tasks: the purpose of the final construction is to provide shelter to a separate object in the scene. The task is also demanding because the obstacles may be elevated far from the floor, and the cost of stickiness essentially prohibits its use.

\textbf{\coveringone{} task}
(Fig.~\ref{fig:tasks}d). Similar to \coveringtwo{}, the agent must build a shelter, but the task is modified to encourage longer term planning: there is a finite supply of movable blocks, the distribution of obstacles is denser, and the cost of stickiness is lower ($-0.5$ per sticky block). It thus incorporates key challenges of the \silhouette{} task (reasoning about which blocks to make sticky), the \reaching{} task (reasoning about precise block layouts), and the \coveringtwo{} task (reasoning about arch-like structures). The limited number of blocks necessitates foresight in planning (e.g. reserving long blocks to cover long obstacles). The reward function and termination criterion are the same as in \coveringtwo{}.

\begin{figure}[t!]
\begin{center}
    \includegraphics[width=0.45\textwidth]{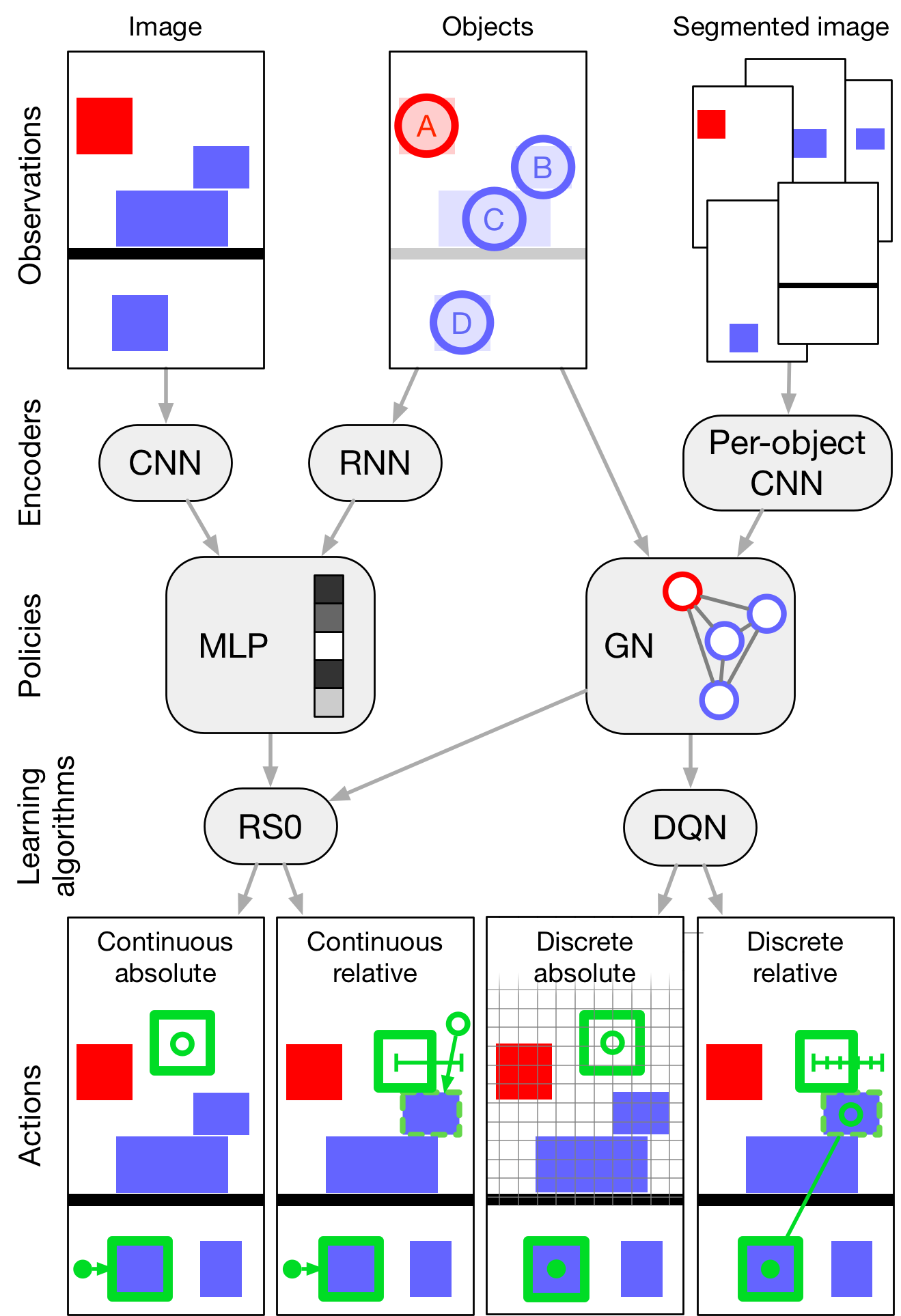}
    \caption{Summary of our construction agents' components. Each agent is defined by an observation format, encoder, policy, learning algorithm, and output action space. We evaluated many of the compatible combinations of these components, as indicated by the grey arrows. For ``continuous absolute'' actions, the agent picks a block from the bottom (solid green circle), and places it in the scene (empty green circle). For ``continuous relative'' actions, the agent picks a block from the bottom (solid green circle), and places it in the scene (empty green circle) with a relative horizontal offset from the nearest block it snaps to. ``Discrete absolute'' actions are similar to continuous absolute actions, except with discrete coordinates. For ``discrete relative'' actions, the agent picks an edge between a block at the bottom (solid green circle) and block in the scene (empty green circle), and a relative horizontal offset.}
    \label{fig:architectures}
\end{center}
\end{figure}

\section{Agents}
\label{sec:agents}

With our suite of construction tasks, we can now tackle the question we posed at the top of the Introduction: what would an agent need to perform complex construction behaviors? 
We expect agents which have explicit structured representations to perform better, due to their capacity for relational reasoning, compositionality, and combinatorial generalization. 
We implement seven construction agents which vary in the degree of structure in their observation types, internal representations, learning algorithms, and action specifications, as summarized in Table~\ref{tbl:architectures} and Fig.~\ref{fig:architectures}.

\begin{table*}
\begin{center}
\begin{tabular}{lcccccccc} 
\toprule
Agent & Observation & Encoder & Policy & Planning & Learning alg. & Action space\\
\midrule
\textbf{RNN-RS0} & Object & RNN & MLP/vector & - & RS0 & Continuous\\
\textbf{CNN-RS0} & Image & CNN & MLP/vector & - & RS0 & Continuous\\
\textbf{GN-RS0} & Object & - & GN/graph & - & RS0 & Continuous\\
\textbf{GN-DQN} & Object & - & GN/graph & - & DQN & Discrete\\
\textbf{GN-DQN-MCTS} & Object & - & GN/graph & MCTS & DQN & Discrete\\
\textbf{CNN-GN-DQN} & Seg. image & Per-object CNN & GN/graph & - & DQN & Discrete\\
\textbf{CNN-GN-DQN-MCTS} & Seg. image & Per-object CNN & GN/graph & MCTS & DQN & Discrete\\
\bottomrule
\end{tabular}
\caption{Full agent architectures. Each component is as described in Sec.~\ref{sec:agents} and also illustrated in Fig.~\ref{fig:architectures}. All agents can be trained with either relative or absolute actions. }
\label{tbl:architectures}
\end{center}
\end{table*}

\subsection{Observation formats}
\label{sec:observations}

Each construction task (Sec.~\ref{sec:tasks}) provides object state and/or image observations.
Both types are important for construction agents to be able to handle: we ultimately want agents that can use symbolic inputs, e.g., the representations in computer-aided design programs, as well as raw sensory inputs, e.g., photographs of a construction site. 

\textbf{Object state}: These observations contain a set of feature vectors that communicate the objects' positions, orientations, sizes, types (e.g., obstacle, movable, sticky, etc.). Contact information between objects is also provided, as well as the order in which objects were placed in the scene (see \supplemental{} Sec.~\ref{appendix_network_architecture}).

\textbf{Image}: Observed images are RGB renderings of the scene, with $(x,y)$ coordinates appended as two extra channels. 

\textbf{Segmented images}: The RGB scene image is combined with a segmentation mask for each object, thus comprising a set of segmented images \citep[similar to][]{janner2019reasoning}.

\subsection{Encoders}
\label{sec:encoders}

\begin{figure}[t!]
\begin{center}
    \includegraphics[width=0.45\textwidth]{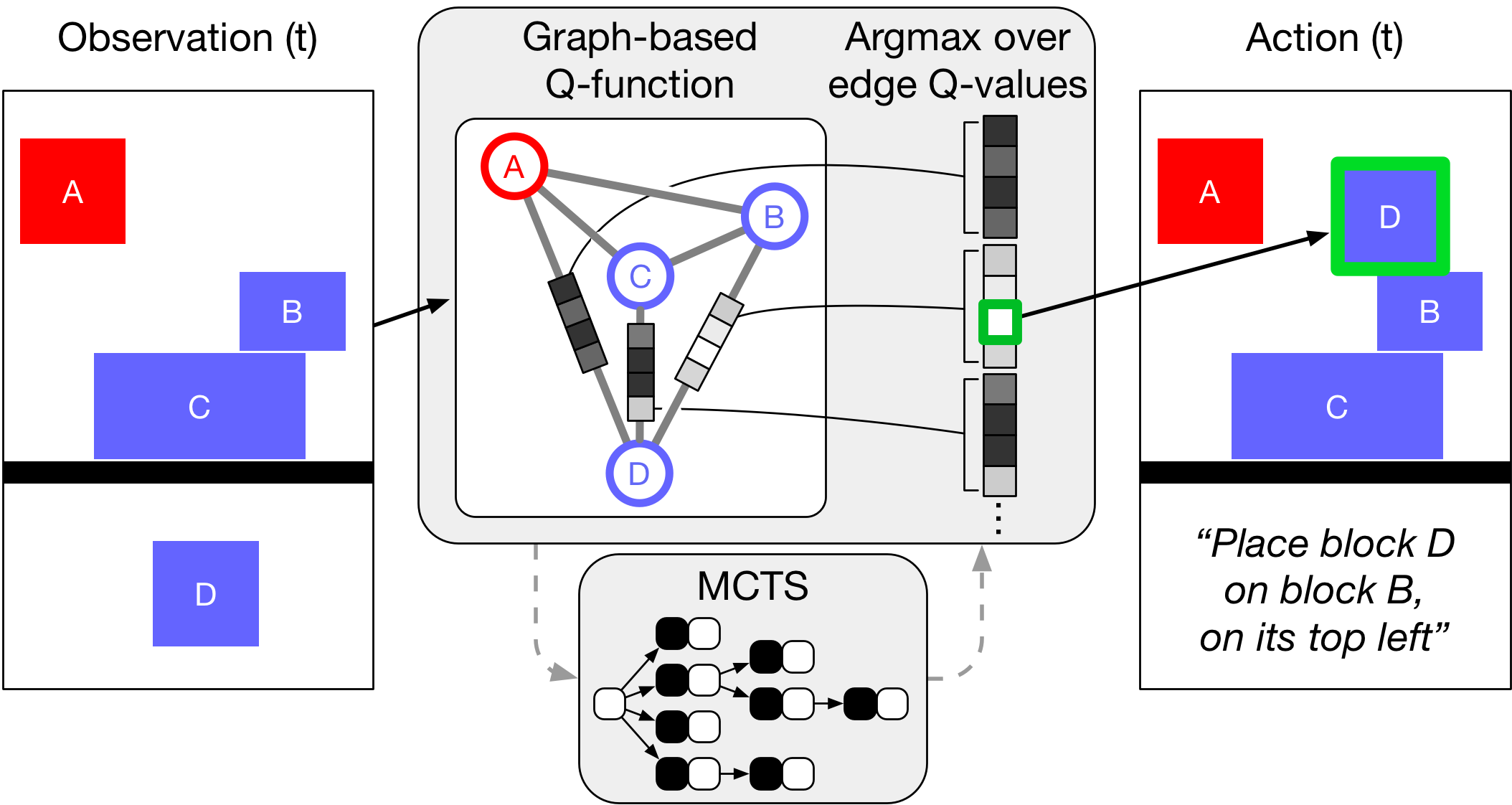}
    \caption{Structure of the GN-DQN agents. The agent takes in a graph-structured representation of the scene where each object corresponds to a node in the graph, and passes this representation through a GN. The GN produces a vector of Q-values for each edge, corresponding to relative actions for picking a block (the start node of the edge) and placing it on another object (the end node of the edge) at a given offset (the edge attribute). To choose actions, the agent takes an $\argmax$ across all edges' Q-values and then converts the edge and offset into $(x,y)$ positions.}
    \label{fig:dqn}
\end{center}
\end{figure}

We use two types of internal representations for computing policies from inputs: fixed-length vectors and directed graphs with attributes.

\textbf{CNN encoder}: The convolutional neural network (CNN) embeds an input image as a vector representation.

\textbf{RNN encoder}: Object state input vectors are processed sequentially with a recurrent neural network (RNN)---a gated recurrent unit (GRU) \citep{cho2014learning}---in the order they were placed in the scene, and the final hidden state vector is used as the embedding.

\textbf{Graph encoder}: To convert a set of state input vectors into a graph, we create a node for each input object, and add edges either between all nodes or a subset of them (see \supplemental{} Sec.~\ref{sec:gn-based-architectures}).

\textbf{Per-object CNN encoder}:
To generate a graph-based representation from images, we first split the input image into segments, and generate new images with only single objects. Each of these are passed to a CNN, and the output vectors are used as nodes in a graph, with edges added as above.

\subsection{Policies}
\label{sec:policies}

\textbf{MLP policy}: Given a vector representation, we obtain a policy using a  multi-layer perceptron (MLP), which outputs actions or Q-values depending on the learning algorithm.

\textbf{GN policy}: Given a graph-based representation from a graph encoder or a per-object CNN, we apply a stack of three graph networks (GN) \cite{battaglia2018relational} arranged in series, where the second net performs some number of recurrent steps, consistent with the ``encode-process-decode'' architecture described in \citet{battaglia2018relational}. Unless otherwise noted, we used three recurrent steps.

\subsection{Actions}
\label{sec:actions}

In typical RL and control settings that involve placing objects, the agent takes \emph{absolute} actions in the frame of reference of the observation \citep[e.g.][]{silver2016mastering,silver2018general,zhang2018composable,ganin2018synthesizing,janner2019reasoning}.
We implement this approach in our ``absolute action'' agents, where, for example, the agent might choose to ``place block D at coordinates $(5.3, 7.2)$''.
However, learning absolute actions scales poorly as the size of the environment grows, because the agent must effectively re-learn its construction policy at every location.

To support learning compositional behaviors which are more invariant to the location in the scene (e.g. stacking one block on top of another), we develop an object-centric alternative to absolute actions which we term \emph{relative} actions.
With relative actions, the agent takes actions in a reference frame relative to one of the objects in the scene.
This is a natural way of expressing actions, and is similar to how humans are thought to choose actions in some behavioral domains \citep{ballard1997deictic,botvinick2004doing}.

The different types of actions are shown at the bottom of Fig.~\ref{fig:architectures}, with details in \supplemental{} Sec.~\ref{sec:actions_impl}.

\textbf{Continuous absolute actions} are 4-tuples $(X, x, y, s)$, where $X$ is a horizontal cursor to choose a block from the available blocks at the bottom of the scene, ``snapping'' to the closest one, $(x, y)$ determines its placement in the scene and the sign of $s$ indicates stickiness (see Sec.~\ref{sec:tasks}).

\textbf{Continuous relative actions} are 5-tuples, $(X, x, y, \Delta x, s)$, where $X$ and $s$ are as before, $(x, y)$ is used to choose a reference block (by snapping to the closest one), and $\Delta x$ determines where to place the objects horizontally relatively to the reference object, the vertical positioning being automatically adjusted.

\begin{figure*}[t!]
\begin{center}
    \includegraphics[width=0.567\textwidth]{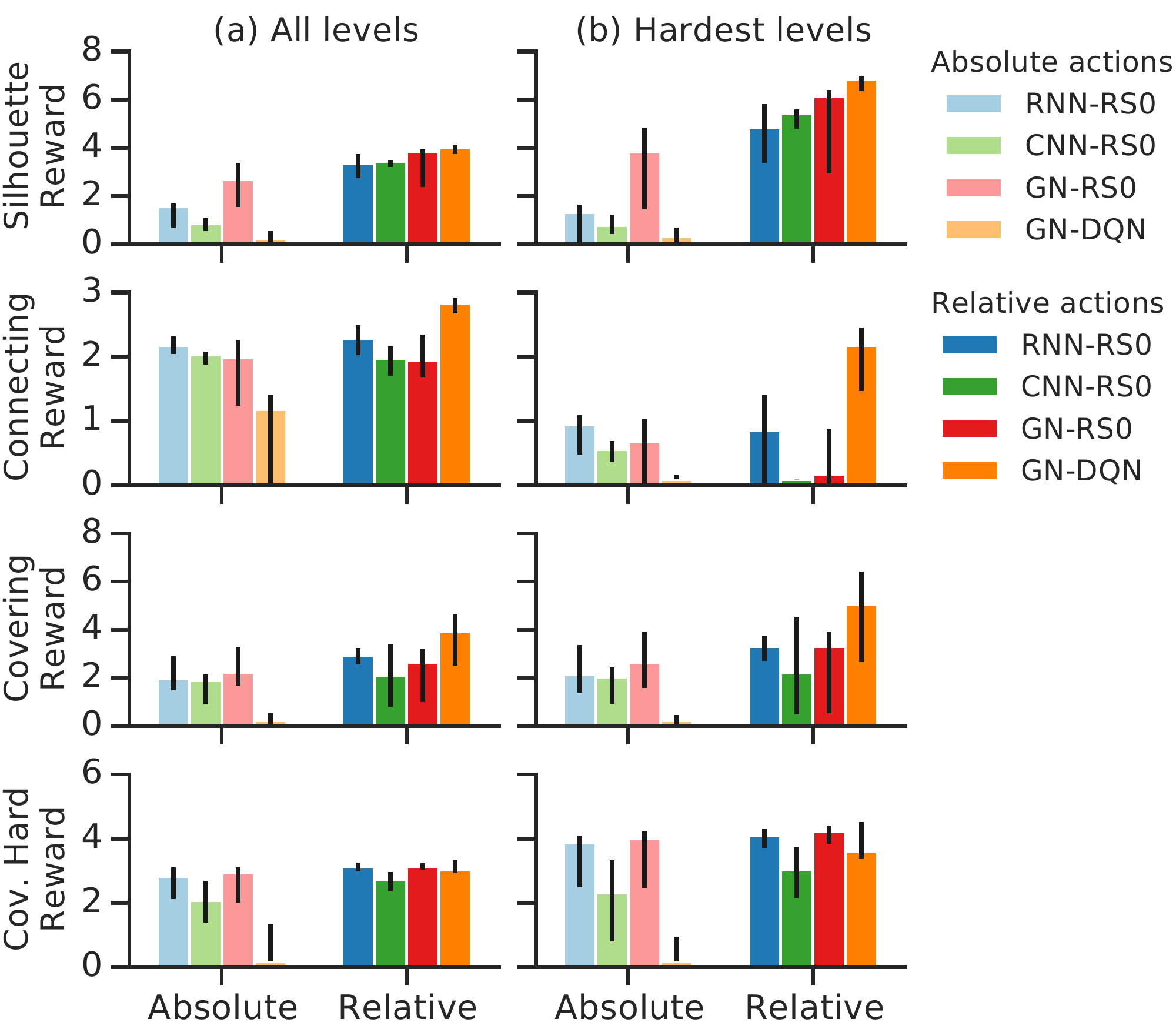}
    \includegraphics[width=0.425\textwidth]{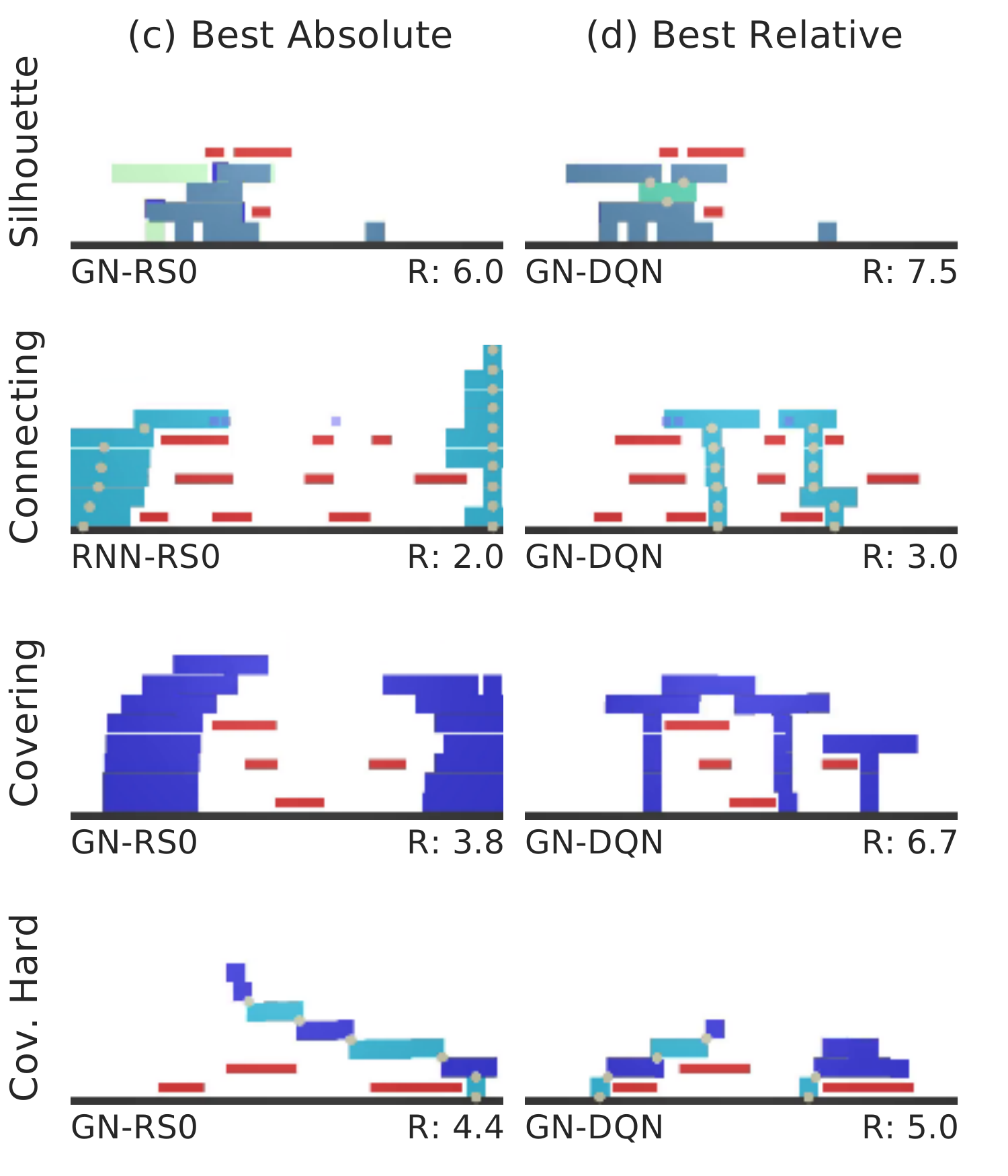}
    \caption{Comparison of absolute and relative actions for model-free agents. (a) Comparison of rewards, averaged across all levels of the curricula. (b) The same as in (a), but for the hardest level of each curricula. (c-d) Qualitative comparison between the best-performing absolute and relative seeds at the hardest curriculum levels in \silhouette{}, \reaching{}, and \coveringtwo{}.}
    \label{fig:absolute-vs-relative}
\end{center}
\end{figure*}

\textbf{Discrete absolute actions} are 4-tuples $(u, i, j, s)$ where $u$ is an index over the available objects, $i, j$ indicate the discrete index at which to place the object in a grid-like 2D discretization of space, and $s$ indicates stickiness.

Absolute actions and continuous relative actions are easily implemented by any agent that outputs a single fixed-length continuous vector, such as that output by an MLP or the global output feature of a GN.

\textbf{Discrete relative actions} are triplets, $(e, i, s)$, where $e := (u, v)$ is an edge in the input graph between the to-be-placed block $u$ and the selected reference block $v$, $i$ is an index over finely discretized horizontal offsets to place the chosen block relatively to the reference block's top surface, and $s$ is as before.

Discrete relative actions are straightforward to implement with a graph-structured internal representation: if the nodes represent objects, then the edges can represent pairwise functions over the objects, such as ``place block D on top of block B'' (see Fig.~\ref{fig:dqn}).

\subsection{Learning algorithms}
\label{sec:learning-algs}

The internal vector and graph representations are used to produce actions either by an explicit policy or a Q-function.

\textbf{RS0 learning algorithm}:
For continuous action outputs, we use an actor-critic learning algorithm that combines retrace with stochastic value gradients (denoted RS0) \citep{munos2016safe,heess2015learning,riedmiller2018learning}.

\textbf{DQN learning algorithm}:
For discrete action outputs, we use Q-learning implemented as a deep Q network (DQN) from \citet{mnih2015human}, with Q-values on the edges, similar to \citet{hamrick2018relational}.
See Sec.~\ref{sec:actions} and Fig.~\ref{fig:dqn}.

\textbf{MCTS}:
Because the DQN agent outputs discrete actions, it is straightforward to combine it with standard planning techniques like Monte-Carlo Tree Search \citep{coulom2006efficient,silver2016mastering} (see Fig.~\ref{fig:dqn}).
We use the base DQN agent as a prior for MCTS, and use MCTS with various budgets
(either only at test time, only during training, or both),
thereby modifying the distribution of experience fed to the learner. As a baseline, we also perform MCTS without the model-free policy prior.
In all results reported in the main text, we use the environment simulator as our model; we also explored using learned models with mixed success (see \supplemental{} Sec.~\ref{subsec_dqn_model_training}).

\begin{figure*}[t!]
\begin{center}
    \raisebox{-0.5\height}{\includegraphics[width=0.395\textwidth]{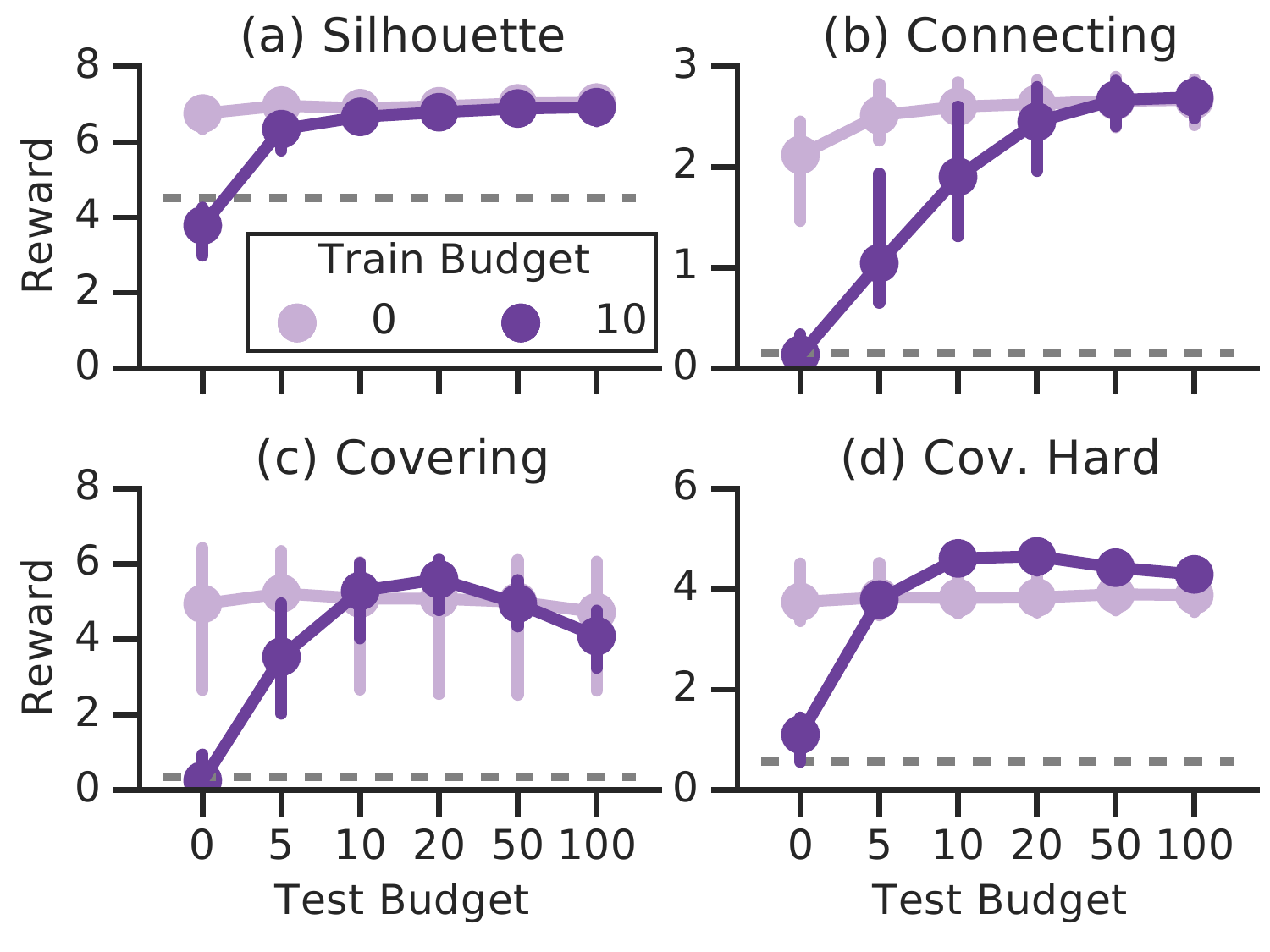}}
    \raisebox{-0.43\height}{\includegraphics[width=0.595\textwidth]{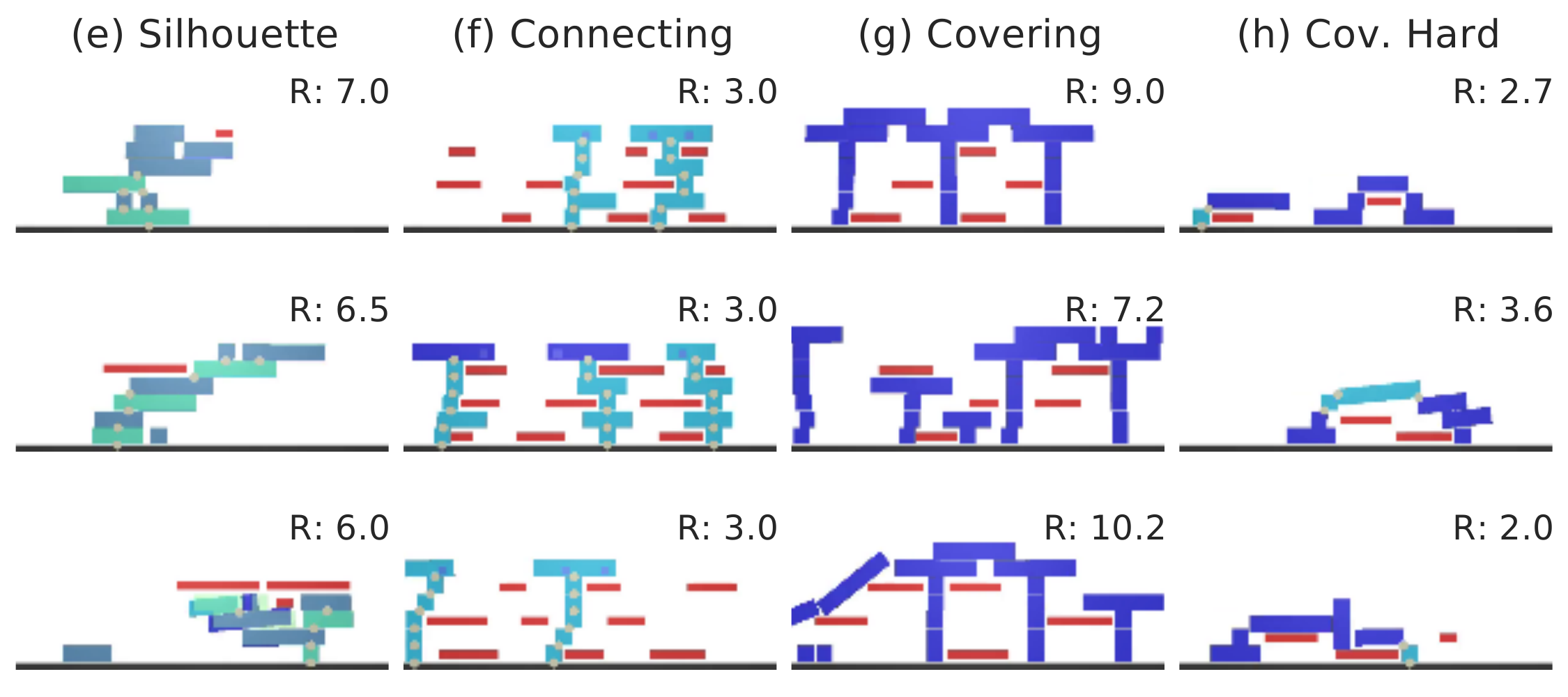}}
    \caption{(a-d) Comparison of different training and testing budgets for the model-based GN-DQN-MCTS agents on the hardest curricula levels. The gray dashed line corresponds to a pure-planning agent with a search budget of 1000. (e-h) Representative structures built by GN-DQN-MCTS agents, chosen from a set of 10 random episodes for each task. The \silhouette{} and \reaching{} agents use training budgets of 0 and test budgets of 50; the \coveringtwo{} agent uses a training budget of 0 and test budget of 5, and the \coveringone{} agent uses a train and test budget of 10. An example of sub-optimal behavior has been chosen for the third row when available. The entire set of random episodes are shown in \supplemental{} Sec.~\ref{sec:example-scenes}.}
    \label{fig:planning}
\end{center}
\end{figure*}

\section{Experiments and results}
\label{sec:results}

We ran experiments to evaluate the effectiveness of different agent architectures (see Table~\ref{tbl:architectures}) on our construction tasks; we also tested several heuristic baselines to estimate bounds on our tasks (see \supplemental{} Sec.~\ref{sec:suppl_tasks}).
We focused on quantifying the effect of structured actions (Sec.~\ref{sec:relative-versus-absolute}), the effect of planning both during training and at decision time (Sec.~\ref{sec:model-free-vs-model-based}), zero-shot generalization performance on larger and more complex scenes (Sec.~\ref{sec:generalization}).
In all experiments, we report results for 10 randomly initialized agents (termed ``seeds'') which were trained until convergence.
Each seed is evaluated on 10,000 scenes, and in all figures we report median performance across seeds as well as errorbars indicating worst and best seed performance.

For efficient training, we found it was important to apply a curriculum which progressively increases the complexity of the task across training episodes. In \silhouette{}, the curriculum increases the number of targets. In \reaching{}, it increases the elevation of the targets. In the \coveringtwo{} tasks, it increases the elevation of the obstacles. Details are available in \supplemental{} Sec.~\ref{sec:suppl_tasks}.
In our analysis, we evaluated each seed on scenes generated either uniformly at random across all difficulty levels, or only at the hardest difficulty level for each task.

\subsection{Relative versus absolute actions}
\label{sec:relative-versus-absolute}

We find that agents which use relative actions consistently outperform those which use absolute actions. Across tasks, almost every relative action agent converges at a similar or higher median performance level (see Fig.~\ref{fig:absolute-vs-relative}a), and the best relative agents achieve up to 1.7 times more reward than the best absolute agents when averaging across all curriculum levels. When considering only the most advanced level, the differences are larger with factors of up to 2.4 (Fig.~\ref{fig:absolute-vs-relative}b).

Fig.~\ref{fig:absolute-vs-relative}c shows examples of the best absolute agents' constructions. These outcomes are qualitatively worse than the best relative agents' (Fig.~\ref{fig:absolute-vs-relative}d).
The absolute agents do not anticipate the long term consequences of their actions as well, sometimes failing to make blocks sticky when necessary, or failing to place required objects at the base of a structure, as in Fig.~\ref{fig:absolute-vs-relative}c's \silhouette{} example.
They also fall into poor local minima, building stacks of blocks on the sides of the scene which fail to reach or cover objects in the center, as in Fig.~\ref{fig:absolute-vs-relative}c's \reaching{} and \coveringtwo{} examples.

By contrast, the best relative agents (which, across all tasks, were GN-DQN) construct more economical solutions (e.g., Fig.~\ref{fig:absolute-vs-relative}d, \reaching{}) and discover richer strategies, such as building arches (Fig.~\ref{fig:absolute-vs-relative}d, \coveringtwo{}).
The GN-DQN agent's superior performance suggests that structured representations and relative, object-centric actions are powerful tools for construction.
Our qualitative results suggest that these tools provide invariance to dimensions such as spatial location, which can be seen in cases where the GN-DQN agent re-uses local block arrangements at different heights and locations, such as the T structures in Fig.~\ref{fig:absolute-vs-relative}g.

Most agents achieve similar levels of performance of \coveringone{}: GN-RS0 has the best median performance, while GN-DQN has the best overall seed.
But inspecting the qualitative results (Fig.~\ref{fig:absolute-vs-relative}), even the best relative agent does not give very strong performance.
Though \coveringone{} involves placing fewer blocks than other tasks because of their limited supply, reasoning about the sequence of blocks to use, which to make sticky, etc. is indeed a challenge, which we will address in the next section with our planning agent.

Interestingly, the GN-RS0 and GN-DQN agents have markedly different performance despite both using the same structured GN policy.
There are a number of subtle differences, but notably, the object-centric information contained in the graph of the GN-RS0 agent must be pooled and passed through the global attribute to produce actions, while the GN-DQN agent \emph{directly} outputs actions via the graph's edges.
This may allow its policy to be more analogous to the actual structure of the problem than the GN-RS0 agent.

The CNN-RS0 agent's performance is generally poorer than the GN-based agents', but the observation formats are also different: the CNN agent must learn to encode images, and it does not receive distinct, parsed objects.
To better control for this, we train a GN-based agent from pixels, labelled CNN-GN-DQN, described in Sec.~\ref{sec:agents}.
The CNN-GN-DQN agent achieves better performance than the CNN-RS0 agent (see \supplemental{} Fig.~\ref{fig:pixels}).
This suggests that parsing images into objects is valuable, and should be investigated further in future work. 

\subsection{Model-based versus model-free}
\label{sec:model-free-vs-model-based}

Generally, complex construction should require longer-term planning, rather than simply reactive decision-making.
Given a limited set of blocks, for example, it may be crucial to reserve certain blocks for roles they uniquely satisfy in the future.
We thus augment our GN-DQN agent with a planning mechanism based on MCTS (see Sec.~\ref{sec:learning-algs}) and evaluate its performance in several conditions, varying the search budget at training and testing time independently (a search budget of 0 corresponds to no planning).

Our results (Fig.~\ref{fig:planning}) show that planning is generally helpful, especially in \reaching{} and \coveringone{}.
In \reaching{}, planning with a train budget of 10 and test budget of 100 improves the agent's median reward from 2.17 to 2.72 on the hardest scenes, or from 72.5\% to 90.6\% of the optimal reward of 3.
In \coveringone{}, planning with a train and test budget of 10 improves the agent's median reward from 3.60 to 4.61. Qualitatively, the planning agent appears to be close to ceiling (Fig.~\ref{fig:planning}h).
Note that a pure-planning agent (Fig.~\ref{fig:planning}a-d, gray dashed line) with a budget of 1000 still performs poorly compared to learned policies, underscoring the difficulty of the combinatorially large search space in construction. 
In \supplemental{} Sec.~\ref{subsec_dqn_tree_expansion}, we discuss of the trade-offs of planning during training, testing, or both.

\begin{figure}[t!]
\begin{center}
    \includegraphics[width=0.48\textwidth]{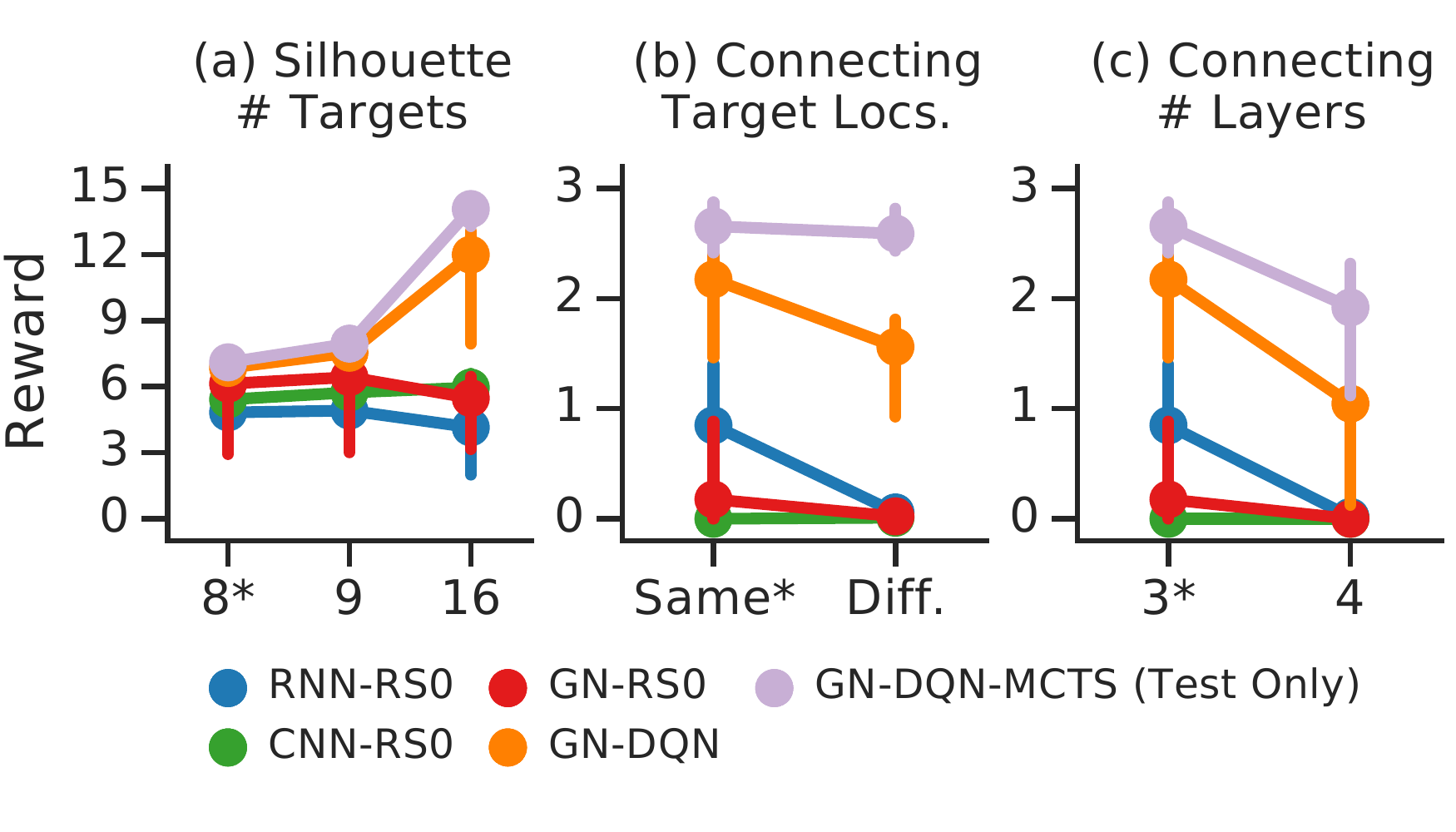}
    \includegraphics[width=0.48\textwidth]{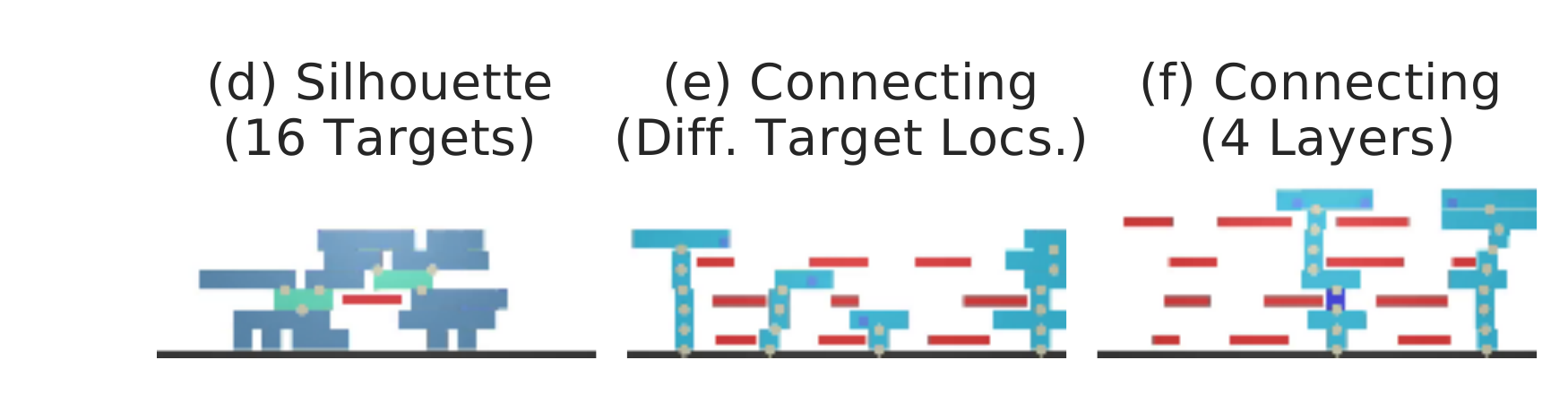}
    \caption{Zero-shot generalization performance of various agents. In all cases, asterisks indicate values seen during training. (a) In \silhouette{}, we varied the number of targets from 8 to 16. (b) In \reaching{}, we first varied the location of the target locations to be either on the same level or on different levels. (c) In \reaching{}, we also varied the number of obstacle layers from three to four. (d-f) Examples of the GN-DQN-MCTS generalizing to new scenes. In each case, the agent has a train budget of 0 and a test budget of 50. (d) Generalization to 16 targets in \silhouette{}. (e) Generalization to multi-level targets in \reaching{}. (f) Generalization to 4 layers of obstacles and higher targets in \reaching{}.}
    \label{fig:generalization-examples}
\end{center}
\end{figure}

\subsection{Generalization}
\label{sec:generalization}

We next ask: how do our agents generalize to conditions beyond those on which they were trained?
In \silhouette{}, our agents only experience 1-8 targets during training, so we test them on 9 and 16 targets.
In \reaching{}, agents always experience targets at the same elevation within one scene during training, so we test them on targets appearing at multiple different levels in the same scene (in one condition) and all at a higher elevation than experienced during training.

We find that the GN-DQN and especially GN-DQN-MCTS agents with relative actions generalize substantially better than others.
In \silhouette{}, the GN-DQN-* agents cover nearly twice as many targets as seen during training, while the other agents' performances plateau or fall off dramatically (Fig.~\ref{fig:generalization-examples}a).
In \reaching{} with targets at multiple different levels, the GN-DQN and GN-DQN-MCTS agents' performances drops only slightly, while other agents' performance drops to near 0 (Fig.~\ref{fig:generalization-examples}b).
With increased numbers of obstacle layers in \reaching{}, both agents' performances drop moderately but remain much better than the less structured agents (Fig.~\ref{fig:generalization-examples}c).
Fig.~\ref{fig:generalization-examples}d-f show the qualitative generalization behavior of the GN-DQN-MCTS agent.
Overall, these generalization results provide evidence that structured agents are more robust to scenarios which are more complex than those in their training distribution. This is likely a consequence of their ability to recognize structural similarity and re-use learned strategies.

\subsection{Iterative relational reasoning}
\label{sec:iterative-relational-reasoning}

Recurrent GNs support iterative relational reasoning by propagating information across the scene graph. We vary the number of recurrent steps in our GN-DQN agent to understand how its relational reasoning capacity affects task performance.
We find that increasing the number of propagation steps from 1 to 3 to 5 generally improves performance (to a point) across all tasks: in \silhouette{}, the median rewards were 3.75, 4.04 and 4.06; in \reaching, 2.49, 2.84, and 2.81; in \coveringtwo{}, 3.41, 3.96, and 4.01; and in \coveringone{}, 2.62, 3.03, and 3.02, respectively.

\section{Discussion}

We introduced a suite of representative physical construction challenges, and a family of RL agents to solve them. Our results suggest that graph-structured representations, model-based planning under model-free search policies, and object-relative actions are valuable ingredients for achieving strong performance and effective generalization. We believe this work is the first to demonstrate agents that can learn rich construction behaviors in complex settings with large numbers of objects (up to 40-50 in some cases), and can satisfy challenging functional objectives that go beyond simply matching a pre-specified goal configuration.

Given the power of object-centric policies, future work should seek to integrate methods for detecting and segmenting objects from computer vision with learned relational reasoning. Regarding planning, this work only scratches the surface, and future efforts should explore learned models and more sophisticated search strategies, perhaps using policy improvement \cite{silver2018general} and gradient-based optimization via differentiable world models \cite{sanchezgonzalez2018graph}.
Finally, procedurally generating problem instances that require complex construction solutions is challenging, and adversarial or other learned approaches may be promising future directions.

Our work is only a first step toward agents which can construct complex, functional structures. However we expect approaches that combine rich structure and powerful learning will be key making fast, durable progress.

\section{Acknowledgements}

We would like to thank Yujia Li, Hanjun Dai, Matt Botvinick, Andrea Tacchetti, Tobias Pfaff, C\'edric Hauteville, Thomas Kipf, Andrew Bolt, Piotr Trochim, Victoria Langston, Nicole Hurley, Tejas Kulkarni, Vlad Mnih, Catalin Ionescu, Tina Zhu, Thomas Hubert, and Vinicius Zambaldi for helpful discussions, input, and feedback on this work.

\bibliography{main}
\bibliographystyle{icml2019}

\appendix
\counterwithin{figure}{section}
\counterwithin{table}{section}
\counterwithin{algorithm}{section}
\onecolumn

\icmltitle{Supplementary Material: \papertitle}

\section{Tasks details}

\subsection{Observation formats}

For each task, the agent could use either an image-based observation, an object-based observation, or a combination of both as a segmentation-masks-based observation.

{Object state observations} are a list of vectors (one for each block), where each vector of size 15 contains information about the corresponding block position ($x, y$), orientation ($\cos(\theta), \sin(\theta)$), size (width, height), linear ($v_x, v_y$) and angular ($v_\theta$) velocities, whether it is sticky or not, and one-hot information about its type (available block, placed block, target or obstacle). The list is ordered by the order under which objects appeared in the scene, but this information is discarded for the graph based agents. Information about which objects are in contact is also provided and is used when constructing the input for the graph based networks (see Sec.~\ref{appendix_network_architecture}).

\textbf{Image observations} start as $128 \times 128$ RGB renders of the scenes and are re-scaled down to $64 \times 64$ by averaging $2\times2$ patches, with the color channels normalized to $[0, 1]$. The $x$ and $y$ coordinate is also supplied for each point in the image and is normalized in the $[-1, 1]$ interval. The re-scaling procedure helps preserve spatial information at a sub-pixel level as color fading at the boundaries between the objects and the background.

\textbf{Segmented images observations} are a list of images, one for each block. They are obtained using a segmentation of the $128 \times 128$ render that maps each pixel to zero or more blocks that may be present at that pixel. Using this segmentation, we build a $128 \times 128$ binary mask for each block, re-scale it down to $64 \times 64$ by averaging $2\times2$ patches, and multiply it with the unsegmented RGB render to obtain per-block renders. We also add the mask as an additional alpha channel to the masked the RGB image, as well as coordinate channels.

\subsection{Full scene generation and reward specifications}
\label{sec:suppl_tasks}

The full rendered scene spans a region a size of 16$\times$16 (meters, unless otherwise indicated).

At the beginning of the episode, the agent has access to 7 available blocks: three small, three medium and one large block (corresponding to respective widths of 0.7, 2.1 and 3.5, all with height 0.7).

The physics simulation is run for 20 seconds after the agent places each block to make sure that the scene is at an equilibrium position before the score is evaluated, and before the agent can place the next block.

\begin{figure*}[!ht]
\begin{center}
    \includegraphics[width=0.9\textwidth]{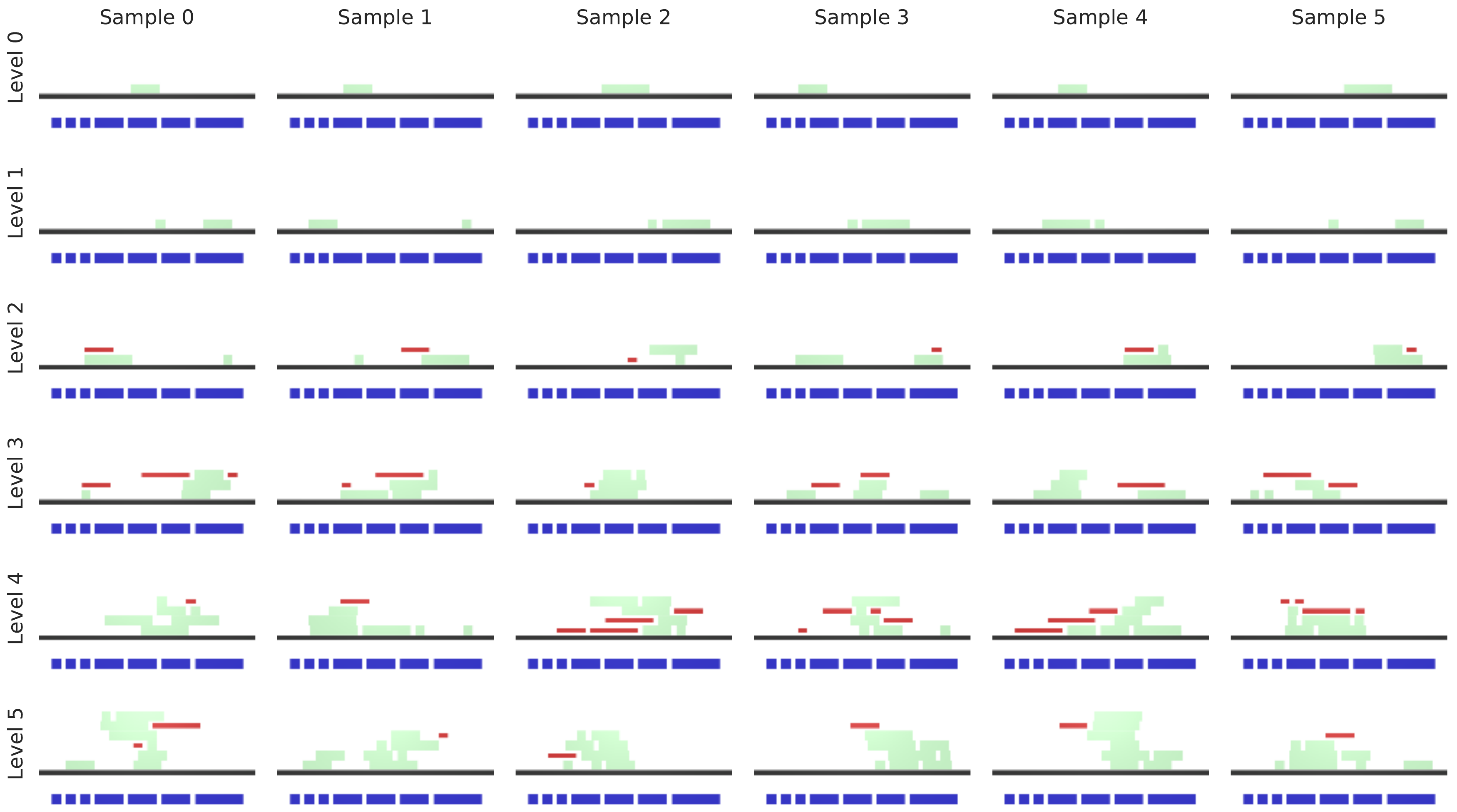}
    \caption{Samples of the hardest scenes for the \silhouette{} task at each level of the curriculum. The $n$-th level of the curriculum consists of scenes uniformly sampled from the rows up to the $n$-th row. Hardest scenes for this task correspond to the $n$-th row.}
    \label{figure_curriculum_silhouette}
\end{center}
\end{figure*}

\textbf{\silhouette{}}:
Each scene is comprised of 1 to 8 targets and 0 to 6 obstacles, arranged in up to 6 layers, with a curriculum over the maximum number of targets, maximum number of obstacles, and number of layers (see Fig.~\ref{figure_curriculum_silhouette}). Levels are generated by (1) tessellating the scene into layers of blocks of the same sizes as the available blocks, with a small separation of 0.35, (2) sequentially (up to the required number of targets) finding the set of target candidates and sampling targets from this set (blocks in the tessellation that are directly on top of the floor or an existing target block) (3) sampling obstacles using a similar procedure. Both obstacles and targets that are further from the floor are sampled with higher probability to favor the generation of harder-to-construct towers and inverted pyramids. The average number of targets is 4.5 on the training distribution, and the number of targets goes up to 8 for the hardest levels. These numbers set an upper bound on the total reward that can be obtained. However, the average reward for an optimal agent is lower than that due to the cost of glue (silhouettes generated using this procedure are not guaranteed to be stable, thus the best possible solution may require glue). We ran a baseline (using the action interface of the Relative GN-DQN agent) consisting of two heuristics for deciding (a) where to place blocks: at target positions, sequentially layer by layer, and from center to the sides and (b) when to use sticky blocks: whenever none of the existing blocks that would touch the new block are sticky and the center of mass of the new block would not be supported by the blocks in the previous layer. This baseline achieves a reward of $3.42$ on the training distribution and $5.27$ on the hardest levels. For comparison our best GN-DQN-MCTS agent seed on this task achieves $4.15$ and $7.18$, respectively.

\begin{figure*}[!ht]
\begin{center}
    \includegraphics[width=0.9\textwidth]{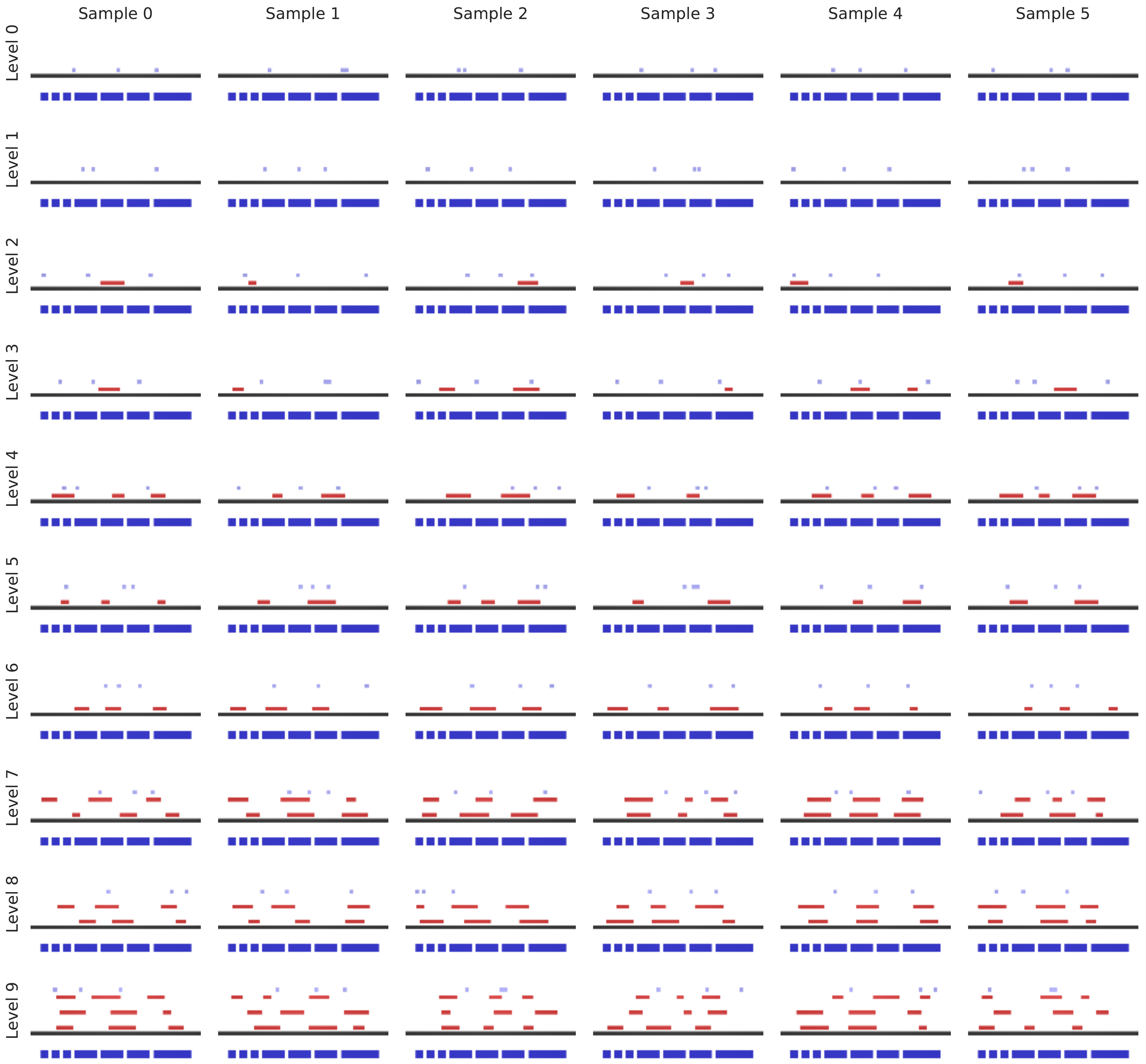}
    \caption{Samples of the hardest scenes for the \reaching{} task at each level of the curriculum. The $n$-th level of the curriculum consists of scenes uniformly sampled from the rows up to the $n$-th row. Hardest scenes for this task correspond to the $n$-th row.}
    \label{figure_curriculum_reaching}
\end{center}
\end{figure*}

\textbf{\reaching{}}:
There are at most three vertical layers of obstacles above the floor and a layer of three targets above the highest obstacles. Each layer consists of up to three obstacles, whose lengths are uniformly and independently sampled from the interval $[0.7, 2.8]$. The layers of obstacles separated by enough distance for one block can be placed between any two layers of obstacles. The curriculum is comprised of scenes fewer obstacle layers, while the number of targets is unchanged (see Fig.~\ref{figure_curriculum_reaching} for examples). Since glue is unpenalized, the maximum reward available to the agent is exactly 3. We expect a heuristic based on path finding would achieve the total reward of 3.

\begin{figure*}[!ht]
\begin{center}
    \includegraphics[width=0.9\textwidth]{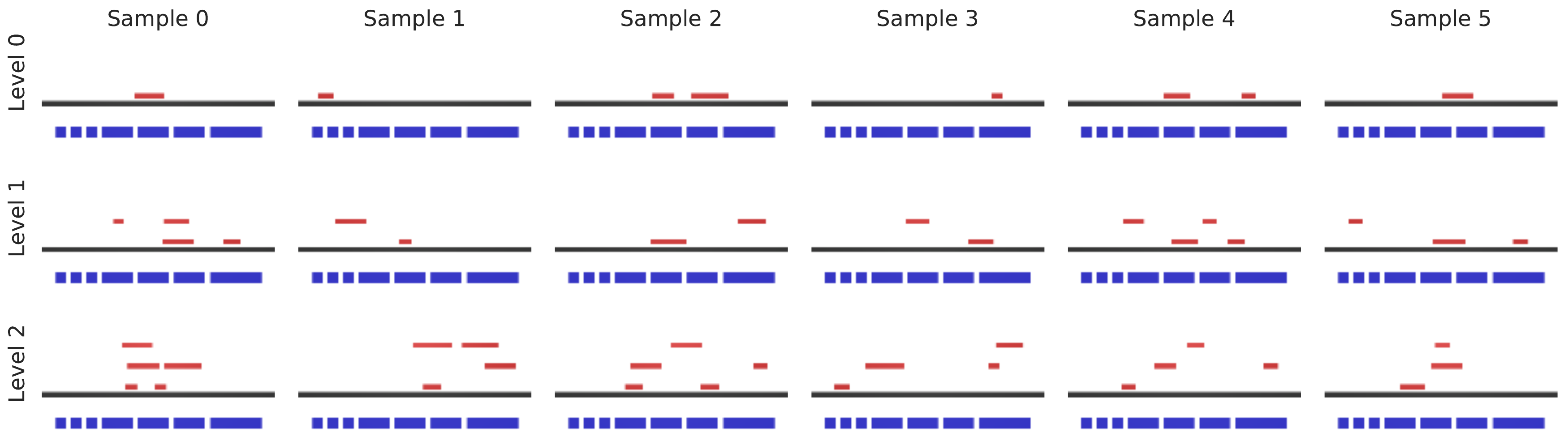}
    \caption{Samples of the hardest scenes for the \coveringtwo{} task at each level of the curriculum. The $n$-th level of the curriculum consists of scenes uniformly sampled from the rows up to the $n$-th row. Hardest scenes for this task correspond to the $n$-th row.}
    \label{figure_curriculum_covering}
\end{center}
\end{figure*}

\textbf{\coveringtwo{}}:
There are at most three vertical layers of obstacles above the floor at any location, and up to 2 obstacles in each layer, with lengths uniformly and independently sampled from the interval $[0.7, 2.8]$. As in \reaching{}, these layers are well separated so that one block can be placed between any two layers of obstacles. The curriculum is comprised of scenes with obstacles only in the first two lower layers (see Fig.~\ref{figure_curriculum_covering}). The total available length to cover is 5.25 on the training distribution and 7.88 for the hardest levels. This provides a tight upper-bound on the maximal reward and the agent could be expected to achieve this. We ran a baseline (using the action interface of the Relative GN-DQN agent) consisting of placing objects layer by layer, prioritizing large blocks and using the following heuristics for (a) odd layers (those with obstacles): place as many blocks as possible in gaps between obstacles, (b) even layers: placing objects to cover the obstacles from the previous layer, then to fill the remaining gaps when possible. This baseline achieves a reward of $3.85$ on the training distribution and $5.31$ on the hardest levels. For comparison our best GN-DQN-MCTS agent seed on this task achieves $4.65$ and $7.18$ respectively.

\begin{figure*}[!ht]
\begin{center}
    \includegraphics[width=0.9\textwidth]{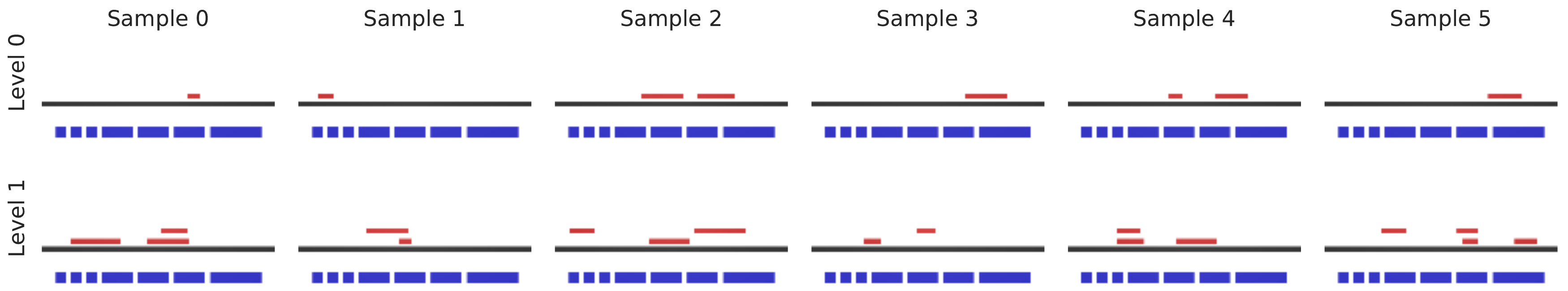}
    \caption{Samples of the hardest scenes for the \coveringone{} task at each level of the curriculum. The $n$-th level of the curriculum consists of scenes sampled from the rows up to the $n$-th row. Hardest scenes for this task correspond to the $n$-th row.}
    \label{figure_curriculum_covering_hard}
\end{center}
\end{figure*}

\textbf{\coveringone{}}:
There are at most two vertical layers of obstacles above the floor at any location, and up to 2 obstacles in each layer, with lengths uniformly and independently sampled in $[0.7, 3.5]$. The curriculum is comprised of scenes with only one layer of obstacles (see Fig.~\ref{figure_curriculum_covering_hard} for examples). The layers of obstacles are closer to each other than in they were in \reaching{} or \coveringone{}. The maximum length that can be covered is 4.2 on the training distribution and 6.3 on the hardest levels, but this only gives a weak upper bound on the possible reward because of the cost of glue and limited supply of blocks. Given the additional complexity of this task, involving resource planning (which blocks to save for later), and balancing the use of sticky blocks (pay price) with the use of arches (use more blocks), we did not find a simple heuristic that could provide a relevant baseline. In particular, note that our heuristic for \coveringtwo{} is not relevant here as it assumes an infinite supply of blocks.

\textbf{Curriculum complexity}:
Curricula were designed to increase in complexity while preserving instances of scenes from previous points in training to avoid catastrophic forgetting. This allows us to make a distinction, for any task and curriculum level, between \emph{Hardest Scenes} (scenes types that are introduced for the first time at the present level) and \emph{All Scenes} (training distribution, including hardest scenes at the current level and lower level scenes).
Additional details about the conditions for advancing through the curricula are given in Sec.~\ref{sec_dqn_details} for the DQN agents and Sec.~\ref{sec_rs0_details} for the RS0 agents.

\section{Implementation details of the action specification.}
\label{sec:actions_impl}

\textbf{Continuous absolute actions} are 4-tuples $(X, x, y, s)$. $X$ is compared to the $x$-coordinates of each of the available blocks and the closest block $c$ is chosen. A new block identical to it is then spawned with its center at location $(x, y)$. The resulting object is sticky if and only if the continuous action $s \in [-1, 1]$ is positive.

\textbf{Discrete absolute actions} are 4-tuples $(u, i, j, s)$. $u \in \{1, 2, \dots,  7\}$ is an index within the set of available blocks to decide which block will be placed on the scene. $i$, $j$ are discrete index to place this block on the scene. Specifically the scene was discretized in height$\times$width using different sizes from $8 \times 64$ to $256 \times 256$, finding the best results for $256 \times 256$ in \silhouette{} and $8 \times 64$ for the other tasks. $s$ is a discrete variable in $\{-1, 1\}$ indicating whether the placed object should be made sticky or not.

\textbf{Continuous relative actions} are 5-tuples, $(X, x, y, \Delta x, s)$. Here $X, s$ have identical meaning to the absolute case, and $c$ is again the object selected by $X$. The object $r$ whose center is closest to $(x, y)$ is then selected as reference. Then, the $x$-coordinate of the placed block $x_p$ is determined by $\Delta x \in [-1, 1]$, such that $x_p=x_r +  \Delta x\left(\frac{w_r+w_c}{2}+\epsilon_x\right)$, where $x_r$ is the $x$-coordinate of the center of $r$, $w_r$ and $w_c$ are the widths of the objects $r$ and $c$, and $\epsilon_x$ is a small offset so that the objects are not touching laterally.

If $r$ is a target object centered at $(x_r, y_r)$, the $y$-coordinate of the center of $c$ will be placed at $y_p = y_r+\epsilon_y$ so that $c$ is vertically overlapping with $r$ (where $\epsilon_y$ is a small offset so that the objects are not perfectly flush).
If $r$ is a solid object, $c$ is placed just above $r$, i.e. $y_p=y_r + \frac{h_r+h_c}{2} + \epsilon_y$, where $h_c$ and $h_r$ are the heights of the objects $c$ and $r$, respectively. If the agent chooses an invalid edge (where $c$ is not an available block, or where $r$ is not a block in the scene), then the episode is terminated with a reward of zero. We use $\epsilon_x=\epsilon_y=0.04$ throughout.

\textbf{Discrete relative actions} are triplets, $(e, i, s)$, where $e := (c, r)$ is an edge in the structured observation graph between the chosen new block and the selected reference block, $i$ is an index over fine discretization of discrete horizontal offsets to place the chosen block relatively to the reference block, and $s$ is as before. If the blocks $c$ are not an available block or that the block $r$ is not a block already in the scene, then the episode is terminated with a reward of 0.

For the $x$ offsets, we use a uniform grid with $n$ bins in the range $[-\frac{w_r + w_c}{2}(1 + \frac{1}{n-3}), \frac{w_r+w_c}{2}(1 + \frac{1}{n-3})]$, where $w_r$ and $w_c$ are as before (this is such that there are exactly $n$ segments in the range $[-\frac{w_r + w_c}{2}, \frac{w_r+w_c}{2}]$). We then pick the $i$-th value in this grid as the relative $x$ position. The $y$ coordinate of the placed block is computed as before, but we also experimented with predicting the relative offset $\Delta y$, and varying the number of discrete offsets $n$ (see Sec.~\ref{sec_dqn_details} for details).

\section{Architecture details}
\label{appendix_network_architecture}

\begin{figure*}[!ht]
\begin{center}
    \includegraphics[width=0.9\textwidth]{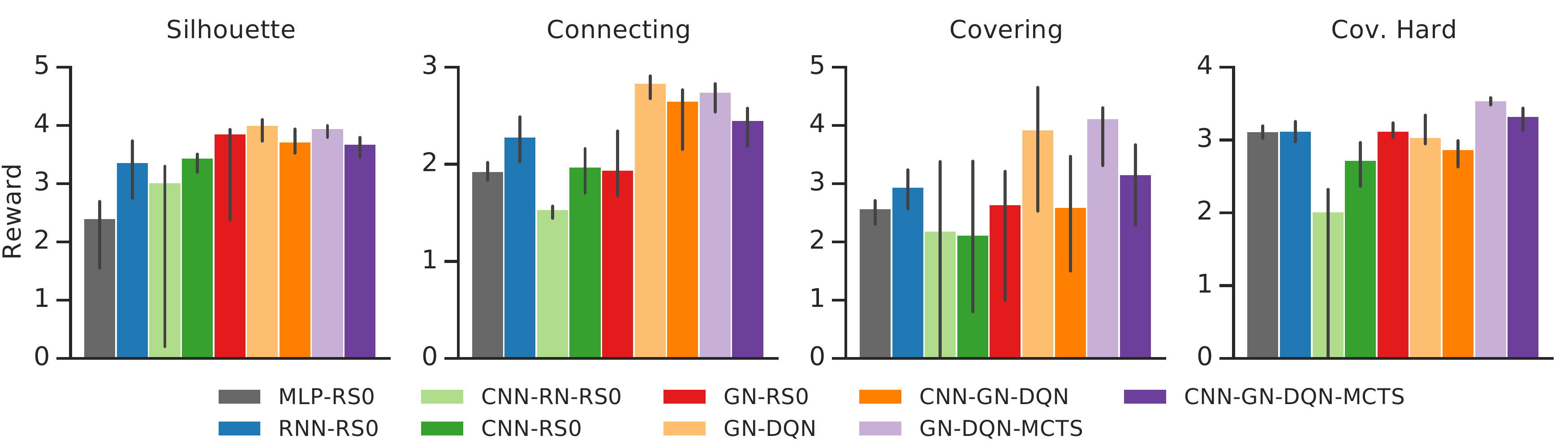}
    \caption{Overall comparison of all agents tested, including MLP and relation network \citep{santoro2017simple} baselines, averaging across curriculum levels. Bars show median performance across seeds, and errorbars are min and max seeds.}
    \label{fig:overall_comparison}
\end{center}
\end{figure*}

\subsection{MLP-based architectures}
\label{sec:mlp-based-architectures}

\textbf{MLP}: The pre-processor of the MLP model consists of concatenating the list of blocks as given by the environment (blocks, available blocks, obstacles, targets) padded with zero blocks (up to the total maximum number of objects in each task with a 1 hot indicator of padding), and normalizing it with a LayerNorm layer into a fixed set 100 features. This fixed size vector is then processed by the core MLP consisting of four hidden layers of $256$ units with ReLU non-linearity, and an output layer to match the required output size.
We found this MLP model to have equal or worse performance to the RNN agent, and thus did not report results on it in the main text; however, Fig.~\ref{fig:overall_comparison} includes results for the MLP agent across the tasks.

\textbf{RNN}: The RNN model pre-processor uses a GRU (hidden size of 256) to sequentially process the objects in the the scene (including padding objects up to a maximum size as described in the MLP). The output of the GRU after processing the last object is then used as input for the core MLP (identical in size to the on described in the MLP model). In some generalization settings, where the total number of objects increased drastically, we found better generalization performance by clipping/ignoring some of the objects in the input, than by allowing the network to process a longer sequence of objects than used at training time.

\textbf{CNN}: The CNN model pre-processor passes the 64$\times$64 input image through a 4-layer convolution network (output channels=[16, 32, 32, 32]) followed by a ReLU activation, a linear layer on the flattened outputs into embedding size of 256, and another ReLU activation. Each layer is comprised of a 2d convolution layer (size=3, stride=1, padding=``same'') and a max pooling layer (size=3, stride=2, padding=``same''). The vector embedding of the image is then processed by and MLP core (identical in size to the on described in the MLP model, except that it uses 3 layers instead of 4).

\textbf{CNN-RN}: We found this CNN-RN model to have equal or worse performance to the vanilla CNN agent, and thus did not report results on it in the main text; however, Fig.~\ref{fig:overall_comparison} includes results for the CNN-RN agent across the tasks.  We use a higher-resolution convolutional feature map, using residual connections to increase depth and ease training.  Each residual block with N channels consists of a N-channel (size=3, stride=1, padding=``same'') convolution and a max pool (size=3, stride=2, padding=``same'').  This is followed by a N-channel convolution (size=3, stride=1, padding=``same''), a ReLU, and another N-channel convolution (size=3, stride=2, padding=``same''), the output of which is added to the max pool output to get the block output.  We apply 3 such blocks with N=[16,32,8].  This gives us a vector of length 8 at every spatial location, to which we apply the standard Relation Net architecture \cite{santoro2017simple}: we contatenate each pairs of these vectors, and feed the length-16 vector into a 2-layer MLP with ReLU activations (64 and 128 units), before applying an average pool operation over all pair representations.  This 128-length vector is a linear layer to produce the final embedding of size 256.

\subsection{GN-based architectures}
\label{sec:gn-based-architectures}

\textbf{Graph pre-processing}:
We use the list of objects or segmentation masks to construct the graphs that are input to the RS0-GN and DQN-GN agents, only discarding the information about the order of appearance of the object in the scene.

For the RS0 agent, we then construct a sparse graph from this set of nodes by connecting (1) available objects to all other objects in the scene; (2) targets and obstacles to all blocks in the scenes; and (3) blocks that are in contact.
The DQN agent takes a fully-connected graph as input but we also experimented with feeding it the sparse representation (see Sec.~\ref{sec_dqn_nrec} for details).

\textbf{GN architecture}:
We use the encode-process-decode architecture described by \citet{battaglia2018relational}. 
comprised of an independent graph encoder, a recurrent graph core with separate MLPs as node, edge, and global functions followed by three GRUs, respectively, and finally as a decoder either a graph network (for the RS0 agent) or graph independent (for the DQN agent).
In symbols, given a graph observation $o$, we process it as
\begin{align*}
  e &= E(o) & o_0 &= e \qquad o_0' = e \\
  o_n &= G([o, o_{n-1}, o_{n-1}'])
  &o_n' &= R(o_{n-1}) \\
  d &= D(o_n') &  & (1 \leq n \leq n_{\textrm{rec}} )
\end{align*}
where $E$ and $D$ are independent graph network (see \citet{battaglia2018relational}), $G$ is a full graph network, and $R$ is a recurrent independent graph network. We use two hidden layers of $64$ units with ReLU non-linearity within all our graph networks.

For this discrete agent, the Q values are finally decoded from $d$ as
$$ q = M([x, d_\textrm{globals}]_{x \in d_\textrm{edges}}),$$
similarly to the approach of \citet{dai2017learning}.

For the RS0 agent we find that having more than 1 recurrent steps in the recurrent graph core did not improve performance so we use a single recurrent step, and disabled the GRU (no longer needed without recurrent steps).

\begin{figure}[t!]
\begin{center}
    \includegraphics[width=0.48\textwidth]{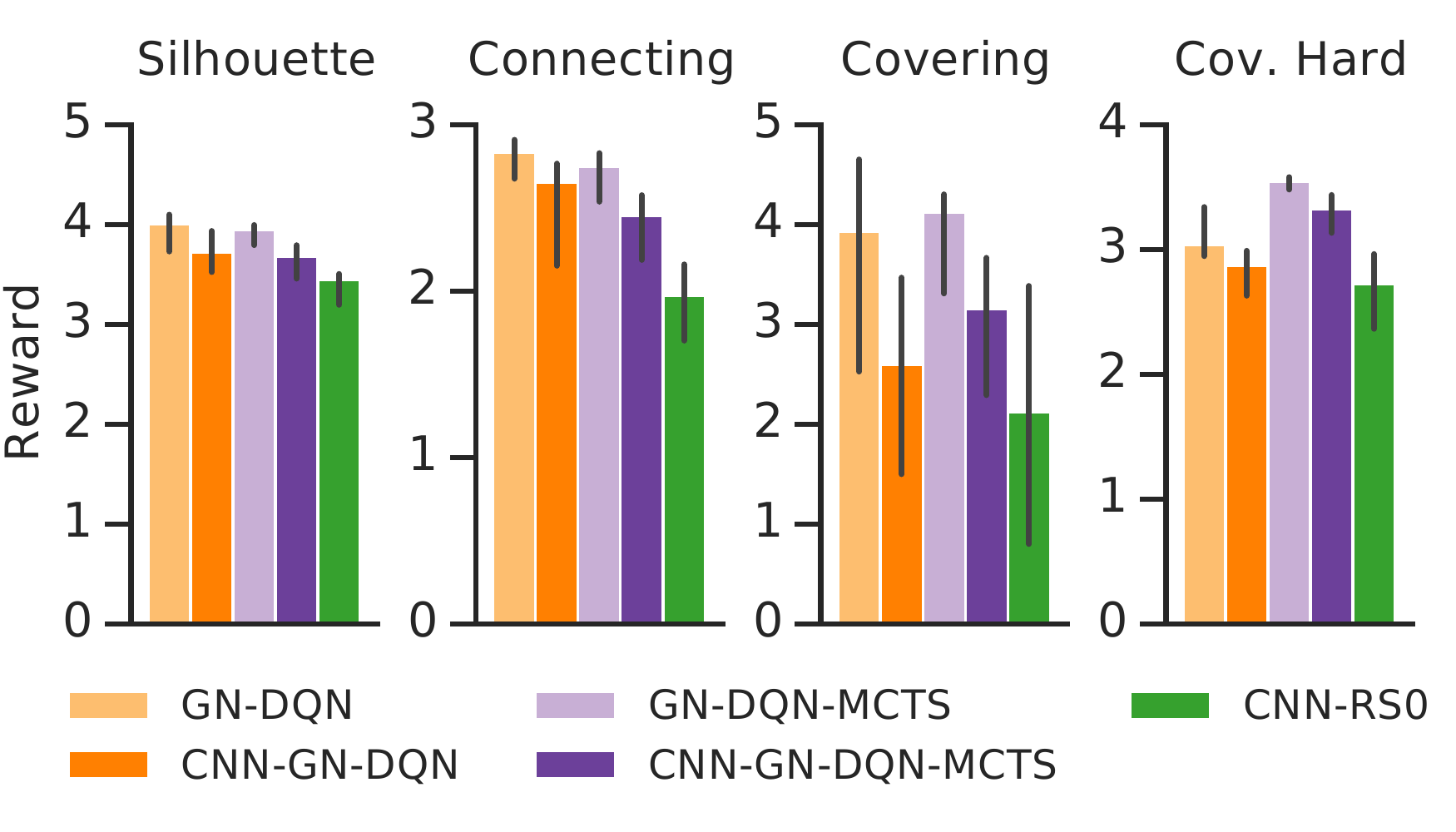}
    \caption{ Comparison of object- and pixel-based agents, averaged across curriculum levels. Bars indicate median performance across seeds, and errorbars are min and max seeds. The CNN-GN-DQN agent augmented with a vision module is able to perform almost as well as its object-based counterparts, and clearly above the CNN-RS0 agent. The same holds for the CNN-GN-DQN agent trained with MCTS, which sometimes even outperforms the object based, model free DQN.}
    \label{fig:pixels}
\end{center}
\end{figure}

\textbf{Segmented images pre-processing}: In the case of the Segmented images observations, each of the nodes in the graph contains an image, which we process independently using a pre-processor similar to that of the CNN model, but smaller (three layers with [8, 16, 8] output channels, followed by two activated linear layers with sizes [64, 32]). This produces a graph with 32 embedded features for each node.

\subsection{Comparison of pixel based methods with objects based methods}

We compare pixel based approaches with object based approaches on Fig.~\ref{fig:pixels}, emphasizing that the graphical networks that take segmented images as input fare closer to their object based graphical counterparts than to raw CNNs, making their usage an exciting avenue for future work.

We also implemented use a CNN followed by a relation network \citep{santoro2017simple}, but it fared worse than the vanilla CNN in all experiments so we excluded it from the main text (see Sec.\ref{sec:mlp-based-architectures} of the \supplemental{}).

\section{Further study on the GN-DQN agent}
\label{sec_dqn_details}

\subsection{General implementation}
We implement a DQN agent with a structured graph input and graph output (roughly similar to \citet{dai2017learning}), but where the Q-function is defined on the edges of the graph. This agent takes a fully-connected graph as input. The actions are decoded from the edges (resp. the global features) of the network's output in the case of the discrete relative (resp. absolute) agent. The learner pulls experience from a replay containing graphs, with a fixed replay ratio of 4. The curriculum over scene difficulty is performed on a fixed, short schedule of $4\times10^4$ learner steps. The main difference with respect to a vanilla DQN is the way we perform $\epsilon$-exploration, which we explain in more detail below.

We use a distributed setup with up to 128 actors (for the largest MCTS budgets) and 1 learner. Our setup is synchronized to keep the replay ratio constant, i.e. the learner only replays each transition a maximum number of times, and conversely actors may wait for the learner to be done processing transitions. This results in an algorithm which has similar learning dynamics to a non-distributed one.

\subsection{$\epsilon$-exploration schedule}

The majority of actions of the discrete agent are invalid, either because they (1) do not correspond to an edge starting from an available block and reaching to an already placed object; (2) because the resulting configuration would have overlapping objects; or (3) because the resulting scene would be unstable.
This has the consequence that doing standard $\epsilon$-exploration strongly reduces the length of an episode (longer episodes are exponentially suppressed), effectively performing more exploration at the beginning of an episode than at its end. To counteract this effect, we use an adaptive $\epsilon$-schedule, where the probability of taking a random action at the $n$-th step of an episode is given by $p_n = \frac{\epsilon}{\textrm{min}(\widehat{L}-n, 1)}$, where $\widehat{L}$ is an empirical estimate of an episodes typical length, and we use $\epsilon = 0.3$ throughout the paper. The final performance is mostly unchanged, but we observe that this makes learning faster and helps with model training (see Sec.~\ref{subsec_dqn_model_training}).

\subsection{Effect of the number of propagation steps and graph connectivity in the graph network}
\label{sec_dqn_nrec}

\begin{figure*}[!t]
\begin{center}
    \includegraphics[width=0.9\textwidth]{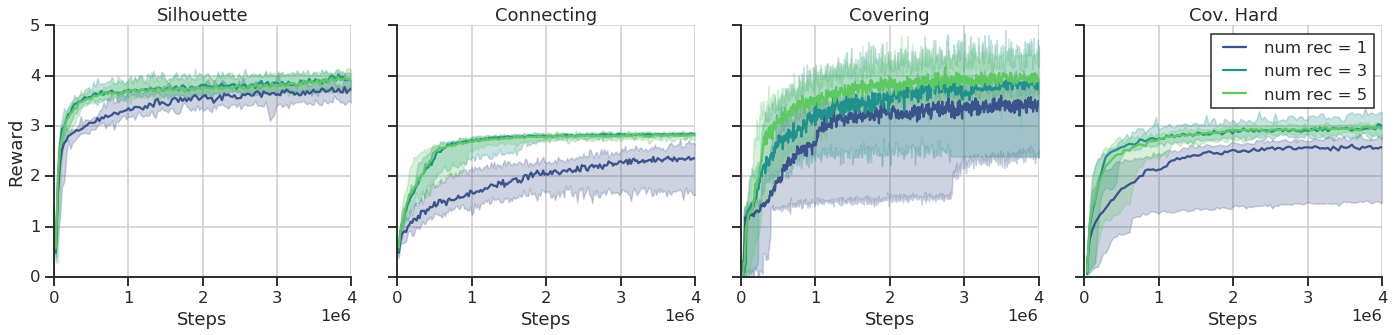}
    \includegraphics[width=0.9\textwidth]{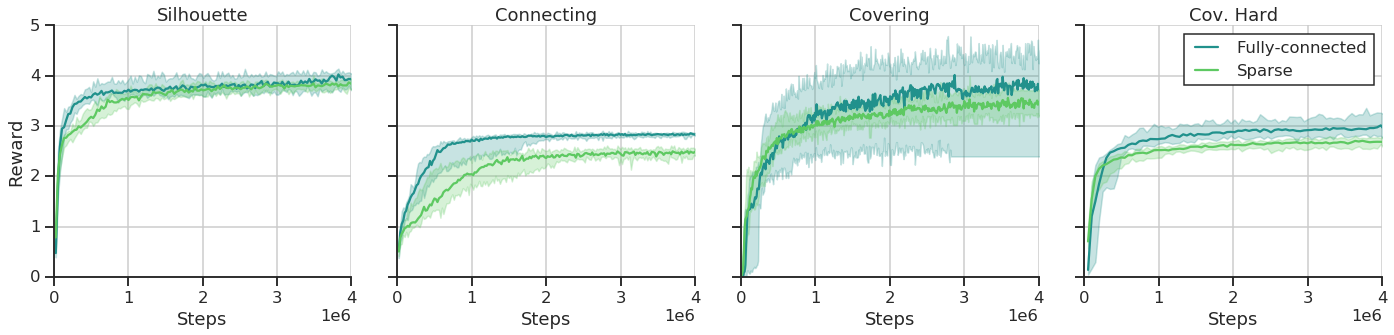}
    \caption{\textbf{Top panel}: Influence of the $n_\textrm{rec}$ parameter on the discrete agent's performance. \textbf{Bottom panel}: Influence of using a sparse graph or the fully-connected graph. Results are qualitatively similar when considering the agents trained with MCTS. In all plots, solid lines are median performance across seeds and shaded regions extend between min and max seeds.}
    \label{figure_dqn_nrec}
\end{center}
\end{figure*}

The results reported elsewhere in this text for the discrete agent were all obtained with $n_\textrm{rec} = 3$ (see Sec.~\ref{sec:gn-based-architectures}) and a fully-connected input graph, but we experimented with varying $n_\textrm{rec}$ and changing the graph connectivity. In Fig.~\ref{figure_dqn_nrec} we show that performance improves with the number of recurrences, but that training is also more unstable, as demonstrated by the wider shaded area around the curve. Empirically, $n_\textrm{rec} = 3$ provides the better compromise between performance and stability.

Those results were all obtained with a fully-connected graph, with a number of edges therefore equal to the number of objects squared. Many of those edges do not however correspond to valid actions or to directly actionable connections, and we experimented with removing those edges from the graph, using the same sparse graph used by the RS0 agent and described in Sec.~\ref{appendix_network_architecture} (note that this graph typically has about 4 times fewer edges than the fully-connected one). What we observe is that this reduces the reasoning capacities of the discrete agent and therefore decreases performance. Augmenting the number of recurrences can partially correct this effect: the best seed with a sparse graph and $n_\textrm{rec} = 7$ can get to the same level of performance as a seed of the fully-connected graph with $n_\textrm{rec} = 3$), but this then happens at the detriment of training stability.

\subsection{Relative actions on both the x and y axes}

\begin{figure*}[!t]
\begin{center}
    \includegraphics[width=0.9\textwidth]{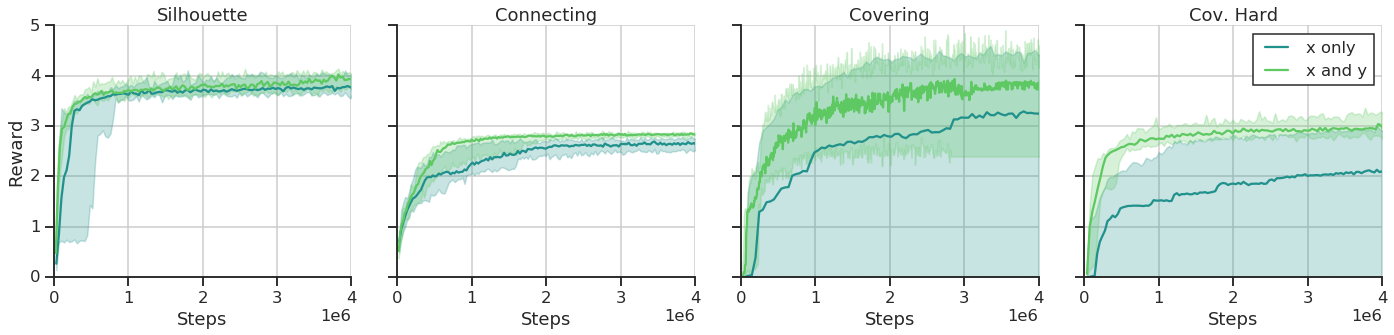}
    \caption{Influence of predicting both the $x$ and $y$ relative coordinates or not. Observe that best performance reaches comparable values in both cases. In all plots, solid lines are median performance across seeds and shaded regions extend between min and max seeds.}
    \label{figure_dqn_y_coordinate}
\end{center}
\end{figure*}

Our discrete relative agents must choose a block to place, an object to use as a reference, and an offset relative to that reference.  Thus far, that offset is only in the $x$-direction, since a small $y$-offset above the reference block is almost always sufficient.  However, what happens if we allow the agent to choose $y$ offset as well?
We observe that this multiplies the size of the action space by the number of discretization points (in our case, $15$), therefore making learning harder. On the other hand, for seeds that manage to start learning, the final performance is equivalent that of the agent which only predicts the relative $x$ position (see Fig.~\ref{figure_dqn_y_coordinate}), despite a number of actions much larger than that of a typical discrete agent, as shown in Table~\ref{table_number_of_actions}.

\subsection{Number of discretization steps}

\begin{figure*}[!t]
\begin{center}
    \includegraphics[width=0.9\textwidth]{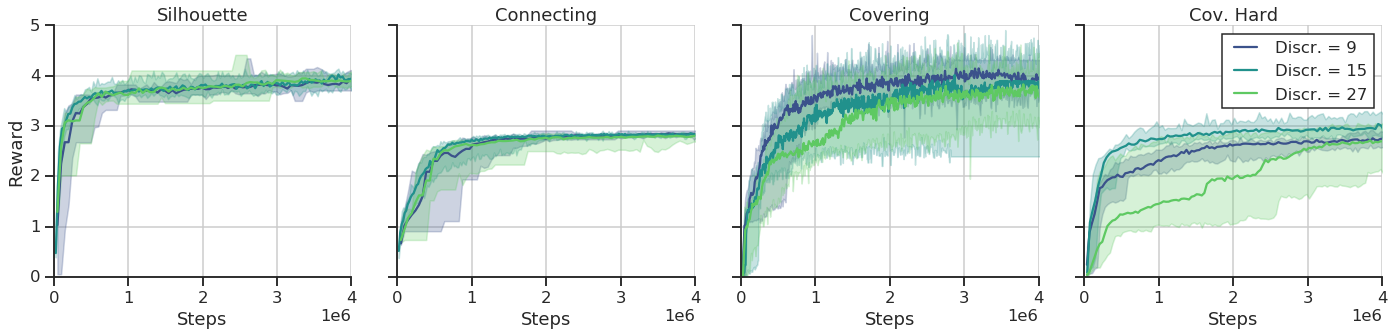}
    \caption{Influence of the number of discretization steps for the relative horizontal positioning of the discrete agent. In all plots, solid lines are median performance across seeds and shaded regions extend between min and max seeds.}
    \label{figure_dqn_discretization}
\end{center}
\end{figure*}

The architecture of the GN-DQN agent naturally represents discrete quantities (i.e., choosing blocks out of a fix set), but using a discrete $x$-offset loses precision over outputting a continuous value. In order to probe the effect of this approximation, we varied the number of discrete locations that the agent is allowed to choose as the second dimension of the action (Fig.~\ref{figure_dqn_discretization}). We observe that a finer discretization of the space allows for slightly better final performance on some problems, but also implies a slower and more unstable learning. Empirically, the 15 steps of discretization used in this paper offers the best compromise. An interesting avenue for further research would be to create an agent that can produce continuous actions attached to a particular edge or vertex of the input graph.

\begin{table}[!ht]
\begin{center}
\begin{tabular}{ |l|c|c|c|c| } 
 \hline
 & Silhouette & Reaching & Covering & Covering Hard \\ 
 \hline
 absolute& $9~10^5$ & $7~10^3$ & $7~10^3$ & $7~10^3$  \\ 
 relative& $2~10^4$ & $2~10^4$ & $2~10^4$ & $3~10^3$ \\ 
 relative ($x, y$) & $2~10^5$ & $3~10^5$ & $3~10^5$  & $5~10^4$ \\ 
 \hline
\end{tabular}
\caption{Typical number of actions for the variations of the tasks, for the discrete agents. The first line reports the number of actions for the best performing discrete absolute agent, the second line for the main discrete relative agent, and the third line for the agents predicting both $x$ and $y$ relative positions.}
\label{table_number_of_actions}
\end{center}
\end{table}

\textbf{Other parameters}:
In all the paper, and unless otherwise specified, we fix the learning rate of the discrete agent to $10^{-4}$ (resp. $2~10^{-4}$) for the model-free (resp. model based) agent and use the Adam optimizer. We use a batch size of 16 and a replay ratio of 4. We perform a linear curriculum over the problem difficulty over a short amount of steps ($4\times10^4$ learner steps). We run all model free agents for $10^7$ learner steps, i.e. approximately $2.5\times10^6$ actor steps. Model based agent are run for up to $4\times10^6$ learner steps ($10^6$ actor steps). Every experiment is run with 10 different seeds.

\section{Further study on the GN-DQN-MCTS agent}
\label{subsec_dqn_tree_expansion}

\subsection{Details of MCTS}

\begin{figure*}[!t]
\begin{center}
    \includegraphics[width=0.9\textwidth]{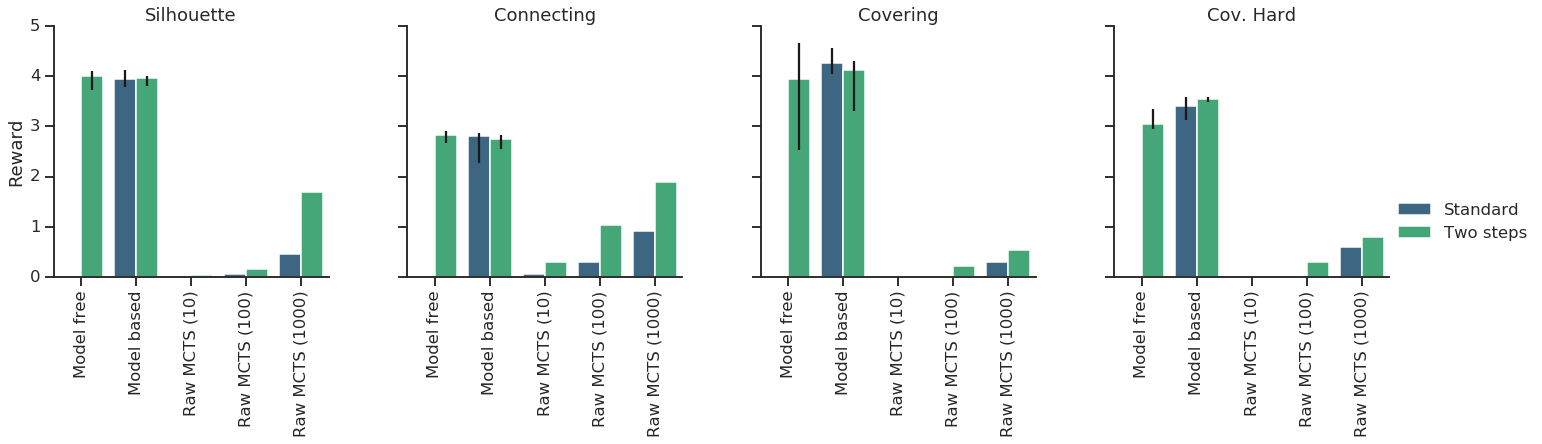}
    \caption{Comparison of the learning based approaches with non-guided MCTS. Model free denotes the relative DQN agent, model based the relative DQN agent with MCTS at training time (budget of 10), Raw MCTS (10) (resp. Raw MCTS (100), Raw MCTS (1000)) denotes a pure MCTS search (without prior on an action value) with a budget of 10 (resp. 100, 1000). For each of the MCTS result, we show the result when performing the search over the actions in two stages (as described in \ref{subsec_dqn_tree_expansion}) or in the usual way. The non-visible bars correspond to zero reward.}
    \label{figure_dqn_mcts_s}
\end{center}
\end{figure*}

The efficiency of Monte-Carlo Tree Search (MCTS) \citep{coulom2006efficient} planning in RL has recently been highlighted in \cite{guo2014atarimcts,silver2016mastering,silver2017mastering,silver2018general}.
Here we combine our DQN agent with MCTS, in the spirit of \citet{sutton91dyna} and \citet{kamyar2018surprising}.
We define a state $s$ in the tree by the sequence of actions that led to it.
In other words, given an episode starting with a configuration $s_0$ and a sequence of actions $(a_0, .., a_t)$, we simply define $s_t:=(a_0, .., a_t)$ (we do not try to regroup states that would correspond to the same observation if coming from different actions sequences). Each node in the tree has a value estimated as
\begin{equation} \label{eq_mcts_q_prior}
    V(s) := \frac{1}{N(s)}\left(\max_{a} Q(s, a) + \sum_{r \in \textrm{rollouts},  s \in r}  \hat{Q}(r, s)\right) , \qquad N(s, a) = 1 +  \sum_{r \in \textrm{rollouts},  s,a \in r} 1 , \qquad N(s)=\sum_a N(s, a)\end{equation}
    (observe that the resulting Monte-Carlo tree has a variable connectivity).
In this expression, $\hat{Q}(r, s)$ is the standard Monte-Carlo return of a rollout after state $s$. The left term $\max_{a} Q(s, a)$ acts as a prior on the value of a node $s$. It is essential to include this term to obtain learning with MCTS, even if using a large budget (see Fig.~\ref{figure_dqn_mcts_s}). We interpret this as being due to the large number of actions stemming from each node and to the fact that many of these actions are actually invalid.

We then perform MCTS exploration by picking the action $a$ that maximizes $ \mathcal{V}(s, a)$, where for common MCTS with UCT exploration \cite{szepsvari2006uct} one would have \begin{equation*}
\begin{split}
\mathcal{V}(s, a) =  V((a_0, \dots, a_t, a)) + c \sqrt{\frac{\ln N(s)}{N((a_0, \dots, a_t, a))}}
\end{split}\end{equation*}
Remembering that the action $a$ can be decomposed as $(\alpha, \beta)$, where $\alpha$ represents an edge index and $\beta$ all the remaining dimensions of the action (relative $x$ placement, use of glue or not, ..), we instead first pick $\alpha$ as the maximizer of  \begin{equation*}
\mathcal{V'}(x, \alpha) := \max_\beta \left [  V((a_0, \dots, a_t, (\alpha, \beta))) \vphantom{\sqrt{\frac{a}{b}}} \right] +  c \sqrt{\frac{\ln N(s)}{\sum_{\beta} N((a_0, \dots, a_t, (\alpha, \beta)))}} ,
\end{equation*}
and then $\beta$ as the maximizer of $\beta \to \mathcal{V}(s, (\alpha, \beta)) $. 
We find this approach to yield slightly better results (see Fig.~\ref{figure_dqn_mcts_s}), and to offer better invariance to changes in the second dimension of the action (e.g. when introducing two dimensional relative placement or changing the number of discretization steps). We use a value of $c=2$ for the UCT constant, and do not find a strong influence of this value on our results.

We then use a transition model to deduce the observation, reward and discount obtained when transitioning from $s$ to $s'$. For the results presented in the main part of this work, we focused on using a perfect transition model, obtained from reseeding the environment every time with the initial state of an episode and reapplying the same sequence of actions. While this is impractical for the large MCTS budgets used in some other works, this provides an upper-bound on the performance that can be obtained with a learnt model and allows to separate hyper-parameters analysis. Also, as we will show in the next paragraph, it is possible to obtain significant gains even when performing the MCTS expansion only at test time.

\subsection{Results with the environment simulator}

We incorporate planning in two ways to our relative discrete agent. In the first variation, we only perform MCTS at test time, using an independently trained Q-network to act as a prior in our MCTS expansion (cf. Eq.(\ref{eq_mcts_q_prior})). We observe that this improves the results on almost all problems but for \emph{Covering}. In particular, in \emph{Reaching}, the fraction of the hardest scenes where the agent does not reach all three targets is decreased by a factor of 4 (from 55\% down to 16\%).

In the second variation, we also perform MCTS at training time: the actor generates trajectories using MCTS expansions using its current Q-function, and the resulting trajectories are then fed to the learner (which does not do any Monte-Carlo sampling). We observe that this second approach yield slightly more stable learning and higher performance on \emph{Covering Hard}, the task that requires the more reasoning (see the last panel of Fig.~\ref{fig:planning}). On the other hand, on other problems, it yields a similar or even decreased performance.

An interesting point to note is that, when training with a perfect simulator, the transfer into the Q-function is very imperfect, as demonstrated by the low value of the left most point on the darker curve of Fig.~\ref{fig:planning}. As it turns out, the agent is relying on the model to select the best action out of the few candidates selected by the Q-function. This may explain why the performance does not necessarily increase when testing with more budget, as the Q-function does not in this case provide a good prior when doing a deeper exploration of the MCTS tree. This is, in essence, also similar to the hypothesis put forward in \cite{kamyar2018surprising}.

\subsection{Results with a learned model}
\label{subsec_dqn_model_training}

Finally, we extend the previous model-based results by performing the MCTS expansion with a learnt model rather than a perfect simulator of the environment.
Using a learnt object-based model was recently put forward as an efficient method for planning in RL \cite{pascanu2017learning, hamrick2017metacontrol} and for predicting physical dynamics \cite{ sanchezgonzalez2018graph, janner2019reasoning}.
Note, however, that none of these approaches have attempted to use MCTS with a graph network-based model.

The model is an operator taking as an input a graph observation and an action, and outputting a new graph observation alongside a reward and discount for the transition: 
\begin{equation*} M: \mathcal{O} \times \mathcal{A} \to \mathcal{O} \times \mathbb{R} \times [0, 1]
\end{equation*}

Given a sequence of observations $o$, actions $a$, rewards $r$ and discounts $\gamma$ belonging to a single episode, we train this model with an unrolled loss
\begin{equation*} \begin{split} L_{n_\textrm{unroll}}((o_{t'})_{t \leq t' \leq t+n_\textrm{unroll}}, (r_{t'}, a_{t'}, \gamma_{t'})_{t \leq t' \leq t+n_\textrm{unroll}-1}) := \sum_{n=0}^{n_\textrm{unroll}-1}  l(M^{(n)}(o_t, (a_{t'})_{t \leq t' < t+n}), r_{t+n}, \gamma_{t+n} )
\end{split} \end{equation*}
where we defined the predicted observation after $n$ steps
\begin{equation*}
    M^{(n)}(o_t, (a_{t'})_{t \leq t' < t+n}) = M(M^{(n-1)}(o_t, (a_{t'})_{t \leq t' < t+n-1}), a_{t+n-1})_0, \qquad M^{(0)}(o_{t}) = o_t
\end{equation*}
and the single step loss
\begin{equation*}
     l((o, r, \gamma), o', r', \gamma') :=  \left\lVert o - o'  \right\rVert_2 + \left\lVert r - r'  \right\rVert_2+ D_{\textrm{KL}}((\gamma, 1-\gamma)|| (\gamma', 1-\gamma')).
\end{equation*}
In practice we varied the number of unrolls $n$ between 1 and 4. The model training is slower with a larger number of unrolls, but it yields more consistent unrolls when used within the MCTS expansion (ideally, the number of unrolls should probably match the typical depth of a MCTS unroll). The model architecture is similar to the one of the main Q-network described in Sec.~\ref{appendix_network_architecture}.

\begin{figure*}[!t]
\begin{center}
    \includegraphics[width=0.9\textwidth]{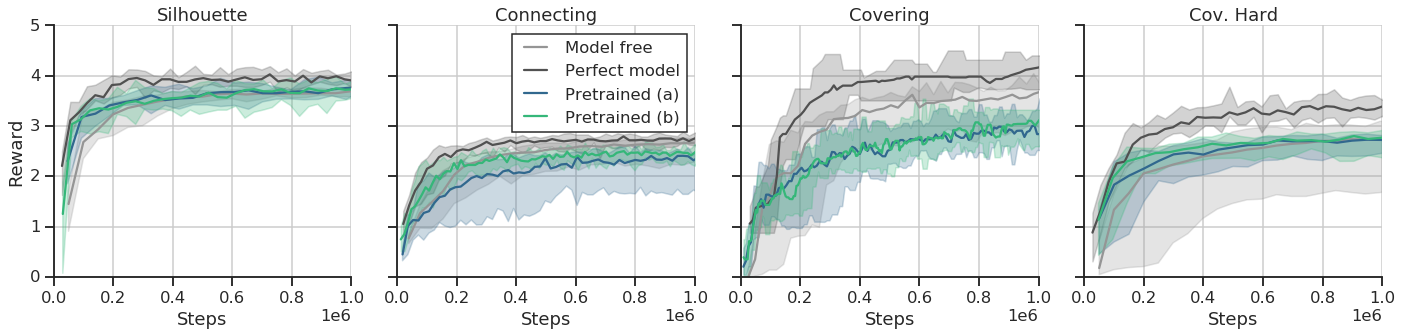}
    \caption{Performance of agents with pre-trained models at small number of steps. \emph{Pretrained (a)} uses a model pre-trained on a model free agent (the continuation of the light grey curve), while \emph{Pretrained (b)} uses a model pre-trained on a model based agent (the continuation of the dark grey curve). Observe how the pre-trained curves match the perfect model curves at short times. All these curve use a same learning rate of $2~10^{-4}$ for fair comparison on short timescales. In all plots, solid lines are median performance across seeds and shaded regions extend between min and max seeds.}
    \label{figure_dqn_mcts_pretraining}
\end{center}
\end{figure*}

\textbf{Model pre-training}:
At first, we experiment with using a pretrained, learnt model to then perform Q-learning with MCTS. The setup is therefore as follows:
\begin{itemize}
\item[(1)] Train an agent model free, or with a perfect environment simulator.
\item[(2)] Train a model on trajectories generated by this agent.
\item[(3)] Train a second agent with the model learnt in (2)
\end{itemize}
We observe in Fig.~\ref{figure_dqn_mcts_pretraining} that this allows to obtain an improved performance at the beginning of training, matching the results obtained with a perfect environment simulator. However, on longer timescales, the performance plateaus and does slightly worse than a model free agent. We interpret this as being due to the rigidity of the model on longer timescales, which is not able to generalize enough to the data distribution that would be required to obtain larger rewards.

\begin{figure*}[!t]
\begin{center}
    \includegraphics[width=0.9\textwidth]{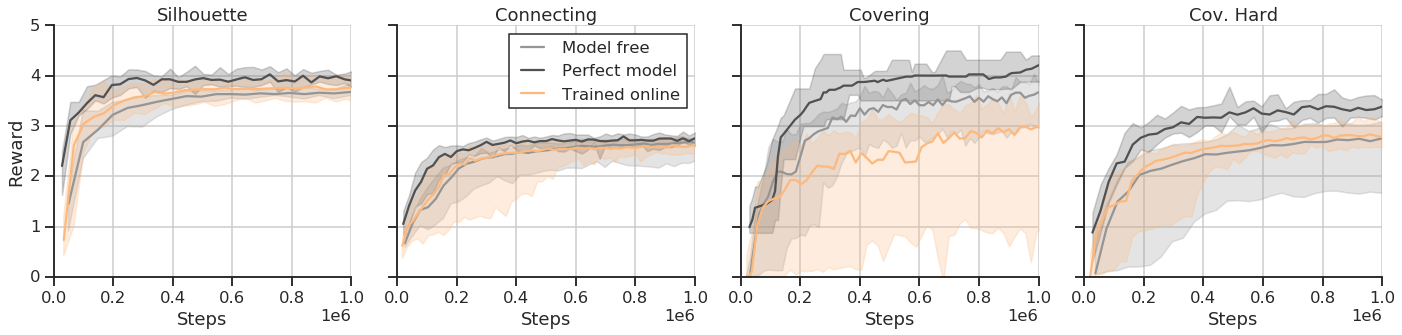}
    \caption{Performance of agents learning a model of the environment at small number of steps. The model based agent use an MCTS budget of 5 that worked best in this setting. All these curve use a same learning rate of $2~10^{-4}$ for fair comparison on short timescales. In all plots, solid lines are median performance across seeds and shaded regions extend between min and max seeds.}
    \label{figure_dqn_mcts_online_model}
\end{center}
\end{figure*}

\textbf{Model learnt online}:
Finally, we try to learn a model online. In this case the agent is trained with a model which is learnt at the same time on trajectories generated by the agent. As shown on Fig.~\ref{figure_dqn_mcts_online_model}, we are able to slightly outperform the model free agent on short timescales in two of the problems (Silhouette and Covering Hard), while the noise introduced by the model is prohibitive again in Covering. On longer timescales, the imperfections of the model make the agent trained with a learnt model converge to the same rewards as the one trained without a model, rather than with a perfect model.

We believe that both in this case and when pre-training the model, understanding how to better train the model so that it generalizes better and yields sharper predictions are important areas of future research, and we see the positive results described here at the beginning of training as a strong motivation to pursue work in this direction.

\section{Model-free continuous agent}
\label{sec_rs0_details}
 
\subsection{RS0} We use a Retraced Stochastic Value Gradients (RS0, \citet{heess2015learning, munos2016safe, riedmiller2018learning}) off-policy agent, with a shared observation pre-processor and independent actor and critic core models (MLP or GN). The critic is conditioned on the actions by concatenating them with the input to the model core (either MLP input features of graph globals). The actor learns a Gaussian distribution over each of the actions by outputting parameters using a linear policy head, conditioned on the last layer of the MLP or output globals of the GN. We use a value of 0.98 for the discount and calculated the retrace loss on sequences of length 5.

\subsection{Exploration}
While the Gaussian policy noise is often sufficient as a source of exploration, due to the highly multi-modal nature of placing objects, we injected additional fixed $\epsilon$ exploration by sampling a continuous action uniformly over the action range with probability $\epsilon$, and sampling from the Gaussian otherwise.  We set $\epsilon=.08$ for tasks with shorter episodes (\silhouette{} and \coveringone{}) and $\epsilon=.03$ otherwise.

\subsection{Dynamic curriculum}

Due to the slower training of RS0 compared to DQN, and the large variance in learning time across the different configurations, we use a dynamic curriculum, only allowing agents to progress through the curriculum once they had achieved a certain performance in the current level.

The criteria for progressing through the curriculum is to obtain at least 50\% of the maximum reward in at least 50\% of the episodes (\silhouette{}, \coveringone{}) or at least 25\% of the maximum reward in at least 25\% of the episodes (\reaching{}, \coveringtwo{}). The threshold values were selected to ensure that the great majority of seeds would reach the maximum level of the curriculum during the allocated experiment time.

To avoid agents from progressing in the curriculum by just solving a particular type of scene, this criteria is applied independently over groups of episodes partitioned based on unique combinations of: number of targets, maximum target height, number of obstacles and maximum obstacle height, and using statistics from the last 200 episodes in each group.
 
\subsection{Distributed setup}

We run every experiment with 10 independent seeds, each of them with 8 actors, 1 learner and 1 FIFO replay (capacity=$10^5$ sequences). Additionally, an evaluation actor with the exploration disabled (and that did not feed data into the replay) is used to generate data to evaluate the dynamic curriculum criteria, and to monitor the overall performance in the task at maximum difficulty.

Due to the off-policy character of the algorithm, we did not set any synchronization between the actors generating the data and the learner obtaining batched of data from the replay. As a consequence, the relative number of actor steps per second and learner steps per second can vary drastically across the different architectures, depending of the relative speed differences between the forward pass (actors), and the backward pass (learner) of the models. Instead, we decided to use a wall-time criteria for terminating our experiments, stopping all experiments after one week of training, or after performance started decreasing. 

\section{Videos}

\begin{table}[hb]
\caption{Links to videos demonstrating constructing behavior for the best agents (as determined by evaluating on 10K episodes) on 10 random episodes. These episodes were taken from the hardest levels of the curriculum for each of the tasks, including Generalization (Gen.) settings where available. For each task, we used the same set of 10 random episodes across all agents to enable easy comparison. All videos are also provided here: \href{https://drive.google.com/drive/folders/1lC9rQTuKYe-XxY0KedK49ymJS0eyicqr?usp=sharing}{https://tinyurl.com/y7wtfen9} .}
\label{table:videos}
\vspace{5mm}
\begin{center}
\begin{small}
\begin{tabular}{p{26.6mm}llll}
\toprule
Task & \parbox[t]{2cm}{Best  \\ absolute agent} & \parbox[t]{2cm}{Best non-GN\\ relative agent}&\parbox[t]{2cm}{Best \\ relative agent} & \parbox[t]{3cm}{Best model-based \\ relative agent} \\
\midrule
\silhouette{} & 
\silhouetteabsoluterszerogn & 
\silhouetterelativerszerornn &
\silhouetterelativedqngn & 
\silhouetterelativedqngnmcts \\

\hspace{0.5cm} Gen. 16 Blocks & 
\silhouetteabsoluterszerogn &
\silhouettegenonerelativerszerocnn &
\silhouettegenonerelativedqngn &
\silhouettegenonerelativedqngnmcts \\

\reaching{} &
\reachingabsoluterszerornn & 
\reachingrelativerszerornn &
\reachingrelativedqngn & 
\reachingrelativedqngnmcts \\

\hspace{0.5cm} Gen. Diff. Locs.&
\reachinggenoneabsoluterszerornn & 
\reachinggenonerelativerszerornn &
\reachinggenonerelativedqngn & 
\reachinggenonerelativedqngnmcts \\

\hspace{0.5cm} Gen. 4 Layers &
\reachinggentwoabsoluterszerornn & 
\reachinggentworelativerszerornn &
\reachinggentworelativedqngn & 
\reachinggentworelativedqngnmcts \\

\coveringtwo{} & 
\coveringtwoabsoluterszerogn & 
\coveringtworelativerszerocnn &
\coveringtworelativedqngn &
\coveringtworelativedqngnmcts \\

\coveringone{} & 
\coveringoneabsoluterszerogn & 
\coveringonerelativerszerornn &
\coveringonerelativedqngn &
\coveringonerelativedqngnmctstrain \\

\bottomrule
\end{tabular}
\end{small}
\end{center}
\vskip -0.1in
\end{table}

\newpage

\section{Further examples of scenes and constructions}
\label{sec:example-scenes}

\begin{figure*}[!ht]
\begin{center}
    \includegraphics[width=0.83\textwidth]{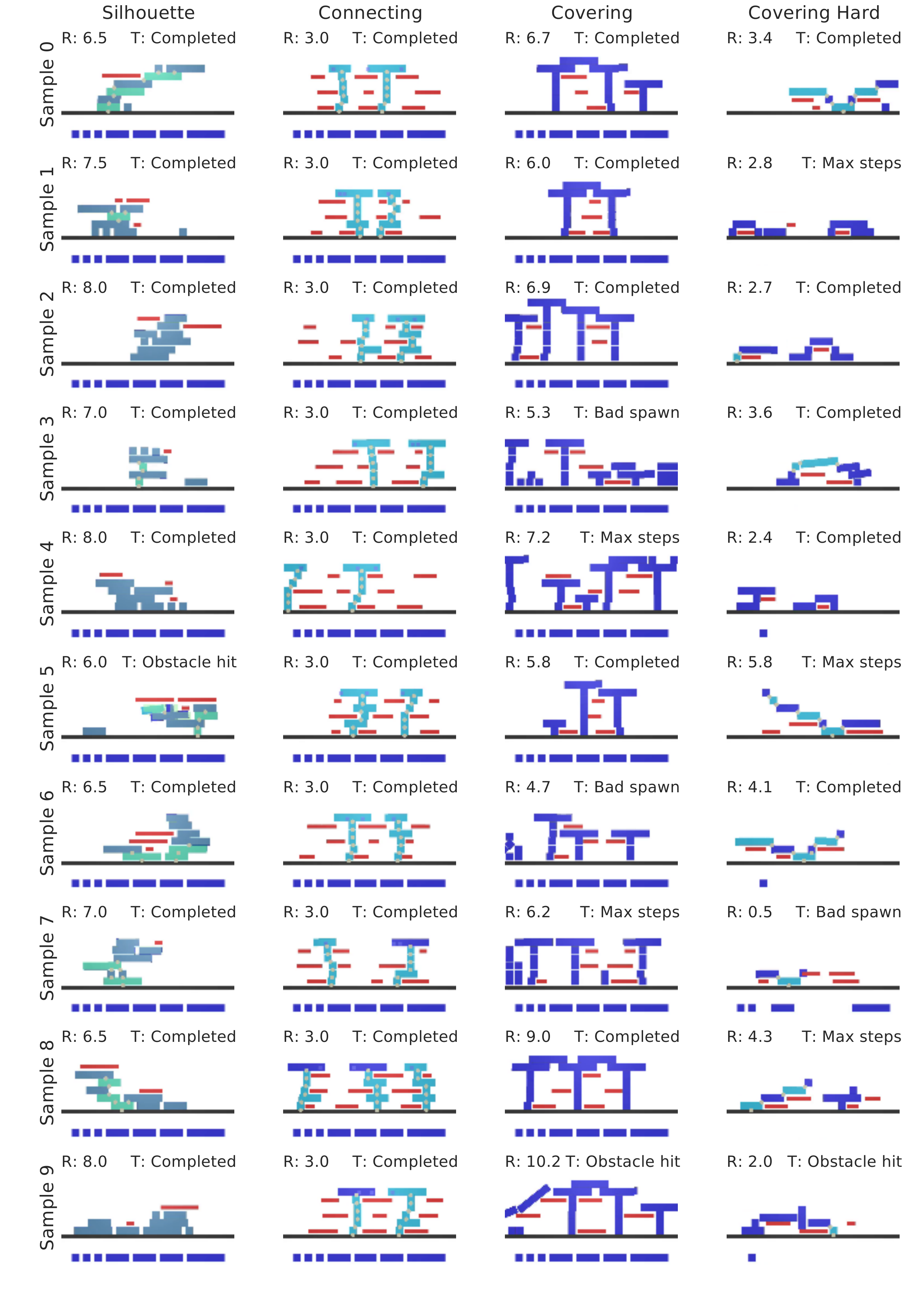}
    \caption{Structures built by the Model-Based agent chosen randomly from the maximum difficulty levels of each task. The total episode reward (R), and termination reason (T) are displayed on top of each scene. Termination reasons are: \emph{Completed} when the agent collects all of the reward in the task, \emph{Max steps} when the agent hits reaches the maximum number of steps or runs out of blocks, \emph{Obstacle hit} when a block hits an obstacle, \emph{Bad spawn} when the agent places a block at a location that makes the new block overlap with an existing block and \emph{Wrong edge} when the DQN agent chooses an edges that is not connecting an available block with a block in the scene.}
    \label{figure_random_model_based_examples}
\end{center}
\end{figure*}

\begin{figure*}[!ht]
\begin{center}
    \includegraphics[width=0.89\textwidth]{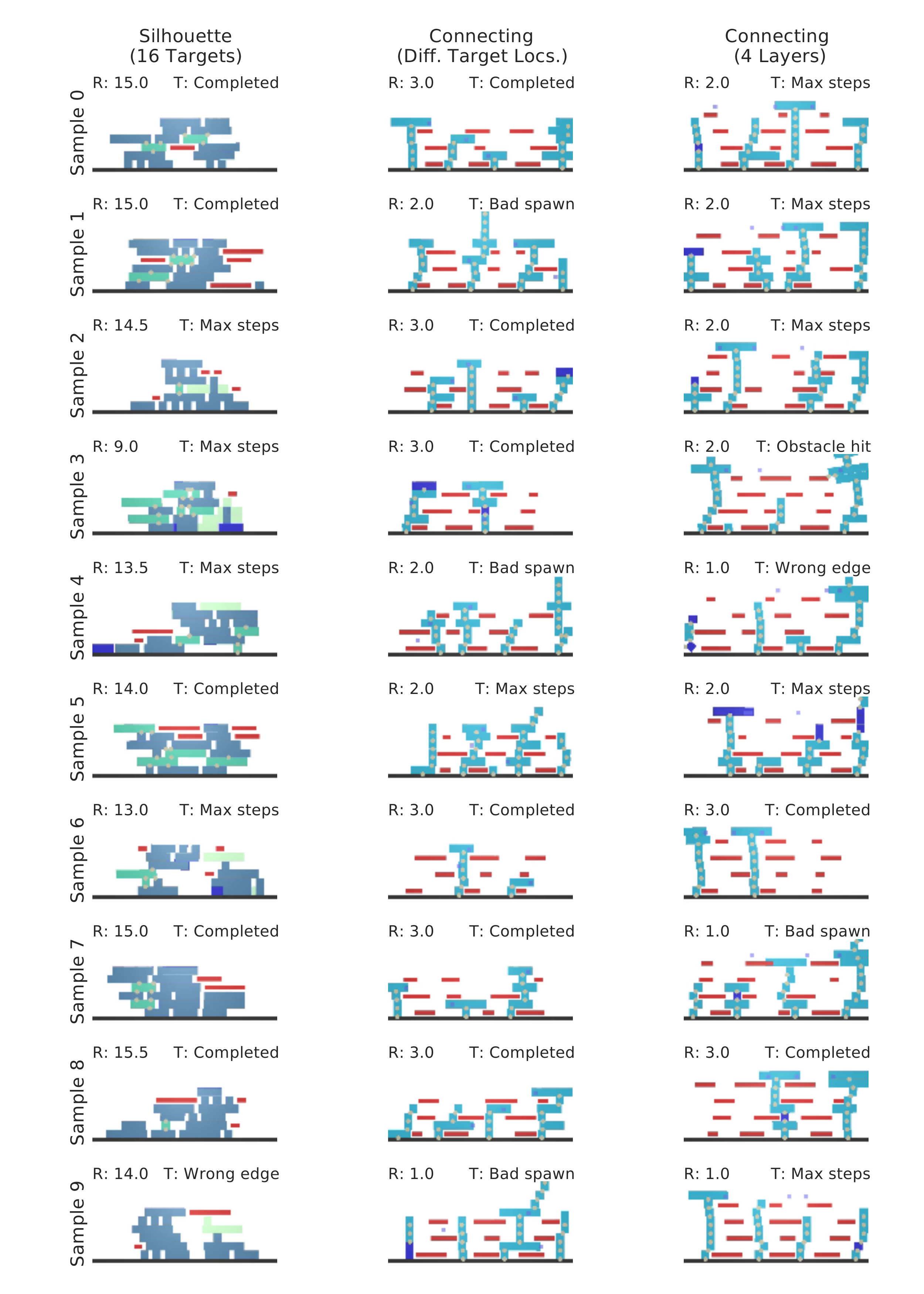}
    \caption{Structures built by the Model-Based agent in generalization settings of the \silhouette{} and \reaching{} tasks. The total episode reward (R), and termination reason (T) are displayed on top of each scene as in Fig.~\ref{figure_random_model_based_examples}.}
    \label{figure_random_model_based_examples_generalization}
\end{center}
\end{figure*}

\end{document}